\documentclass[10pt,twocolumn,letterpaper]{article}

\usepackage{cvpr}              
\definecolor{cvprblue}{rgb}{0.21,0.49,0.74}
\usepackage[pagebackref,breaklinks,colorlinks,allcolors=cvprblue]{hyperref}

\usepackage{multirow}  
\RequirePackage[dvipsnames]{xcolor}
\RequirePackage[table]{xcolor}
\usepackage{amsmath}
\usepackage{amssymb}
\usepackage{graphicx}

\def\paperID{10791} 
\def\confName{CVPR}
\def\confYear{2026}

\title{WikiCLIP: An Efficient Contrastive Baseline for Open-domain Visual Entity Recognition}

\author {
    Shan Ning\textsuperscript{\rm 1,\rm 3},
    Longtian Qiu\textsuperscript{\rm 1},
    Jiaxuan Sun\textsuperscript{\rm 1},
    Xuming He\textsuperscript{\rm 1,\rm 2} \\
    \textsuperscript{\rm 1}ShanghaiTech University, Shanghai, China\\
    \textsuperscript{\rm 2}Shanghai Engineering Research Center of Intelligent Vision and Imaging\\
    \textsuperscript{\rm 3}Lingang Laboratory, Shanghai, China\\
    \{ningshan2022, qiult, sunjx2022, hexm\}@shanghaitech.edu.cn
}

\begin{document}
\maketitle
\begin{abstract}

Open-domain visual entity recognition (VER) seeks to associate images with entities in encyclopedic knowledge bases such as Wikipedia.
Recent generative methods tailored for VER demonstrate strong performance but incur high computational costs, limiting their scalability and practical deployment.
In this work, we revisit the contrastive paradigm for VER and introduce WikiCLIP, a simple yet effective framework that establishes a strong and efficient baseline for open-domain VER.
WikiCLIP leverages large language model embeddings as knowledge-rich entity representations and enhances them with a Vision-Guided Knowledge Adaptor (VGKA) that aligns textual semantics with visual cues at the patch level. To further encourage fine-grained discrimination, a Hard Negative Synthesis Mechanism generates visually similar but semantically distinct negatives during training.
Experimental results on popular open-domain VER benchmarks, such as OVEN, demonstrate that WikiCLIP significantly outperforms strong baselines. Specifically, WikiCLIP achieves a 16\% improvement on the challenging OVEN unseen set, while reducing inference latency by nearly 100 times compared with the leading generative model, AutoVER. The project page is available at \href{https://artanic30.github.io/project_pages/WikiCLIP/}{github/WikiCLIP}.

\end{abstract}

\begin{figure}[ht]
    \centering
     \includegraphics[width=0.9\linewidth]{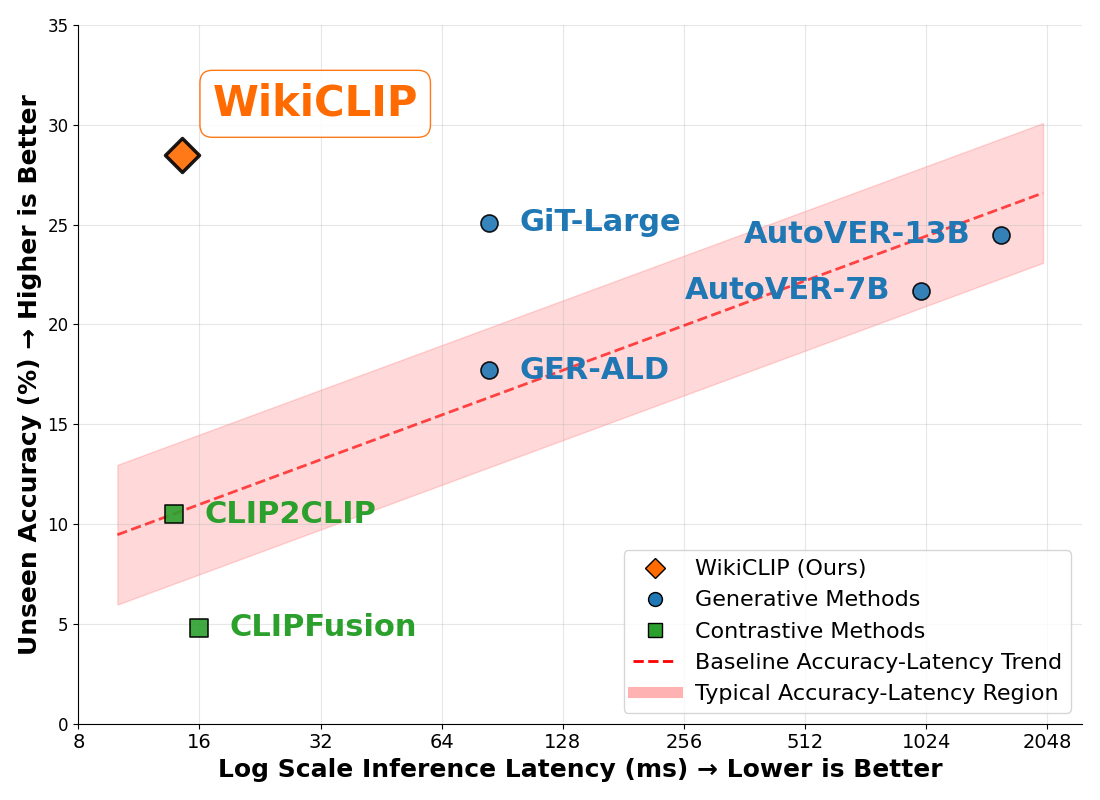}   
    \caption{\textbf{Performance comparison.} We compare WikiCLIP with strong baselines in terms of inference latency and generalization ability, as measured by unseen accuracy. As shown in the figure, WikiCLIP delivers state-of-the-art performance in OVEN unseen accuracy while maintaining low inference latency.
    }
    \label{fig:teaser2}
    \vspace{-6mm}
\end{figure}


\section{Introduction}
Open-domain Visual Entity Recognition (VER) aims to identify specific named entities appearing in an image, where the entity space is drawn from encyclopedic knowledge sources such as Wikipedia. 
VER serves as a critical component in various real-world applications, including information-seeking Visual Question Answering (VQA)~\citep{infoseek,EVQA,ning2026wiki}, animal species recognition~\citep{Wah2011TheCB,Horn2017TheIS,Khosla2012NovelDF}, and news content understanding~\citep{Biten_Gomez_Rusinol_Karatzas_2019,Fu_Wang_Yang_2021,Liu_Wang_Wang_Ordonez_2021}.
Despite rapid advances in multimodal large language models (MLLMs), recent studies~\citep{autover} reveal that VER remains highly challenging. It demands reasoning over fine-grained encyclopedic knowledge and recognizing entities across an extremely large and long-tailed category space—often encompassing millions of candidates. 
While modern MLLMs~\citep{openai2023gpt4v} excel at open-ended reasoning, they still struggle to encode long-tail factual knowledge and to ground it precisely in visual evidence~\citep{autover}. This underscores the continuing need for dedicated research on VER models.

Recent works on open-domain VER~\citep{GERALD,autover,REW} suggest that generative paradigms-which translate query images into text and then perform text-based entity matching against encyclopedic sources-currently outperform contrastive approaches (e.g., CLIP-based methods~\citep{infoseek}). 
However, generative methods suffer from several critical limitations:
(1) High inference latency – autoregressive decoding requires sequential token generation, causing significant computational overhead compared to parallelizable contrastive encoders.
(2) Limited generalization to unseen entities – Generative VER models often fail to recognize entities not observed during VER training.
(3) High computational cost – They typically rely on massive architectures and large-scale paired datasets; for example, AutoVER~\citep{autover} uses 13B parameters, and REW~\citep{REW} trains on 47M image–text pairs.
These drawbacks become especially pronounced when generative VER models are integrated as intermediate modules within larger pipelines, where they can cause slow inference, reduced adaptability, and cumulative error propagation in downstream tasks.

In this work, we revisit the contrastive paradigm for VER and argue that it remains a powerful yet underexplored alternative to generative approaches. 
We propose WikiCLIP, a simple and efficient framework that establishes a strong contrastive baseline for open-domain VER.
Our key insight is that LLM embeddings can encode rich encyclopedic semantics when provided with textual descriptions. By guiding these representations with fine-grained visual cues, we can extract discriminative, entity-level embeddings through lightweight contrastive training. 
WikiCLIP thus combines the generalization ability of LLM-based representations with the efficiency and scalability of contrastive learning, offering a practical and robust solution for open-domain VER.

Specifically, we instantiate WikiCLIP with a \textit{Vision-Guided Knowledge Adaptor (VGKA)} module and a \textit{Hard Negative Synthesis} in training. 
The VGKA produces knowledge-aware entity representations for efficient retrieval-based matching against image queries. 
Given an entity's text-image pair, the model first employs a large language model to derive a semantic-rich text representation. Crucially, this representation is then aligned with and filtered by patch-level visual features extracted using a CLIP encoder~\citep{CLIP}, enabling the VGKA to focus on entity-relevant semantics within lengthy, complex texts while suppressing irrelevant information. 
The VGKA is optimized using a contrastive learning objective that aligns knowledge-enhanced entity representations with corresponding query visual counterparts.
To further strengthen this selective alignment, we develop a Hard Negative Synthesis mechanism to generate challenging negatives by swapping entity texts with descriptions of visually similar entities, forcing the model to capture fine-grained textual distinctions that define entity identity.

We evaluate WIKICLIP on three open-domain Visual Entity Recognition benchmarks: OVEN~\citep{OVEN}, INFOSEEK~\citep{infoseek}, and EVQA~\citep{EVQA}.
WIKICLIP achieves competitive performance compared to strong generative baselines while being significantly faster — for instance, \textit{14.49 ms} vs. \textit{1569 ms} when compared with the state-of-the-art AutoVER.
Moreover, WIKICLIP demonstrates superior generalization, achieving 28.5\% accuracy on unseen entities in OVEN, surpassing the previous best result of 25.1\%. Our key contributions are:

    
    
    

\begin{itemize}
    \item We propose WikiCLIP, a simple and efficient framework that establishes a strong contrastive baseline for open-domain VER.
    \item We introduce a Vision-Guided Knowledge Adaptor and a Hard Negative Synthesis Mechanism, which together enable fine-grained, entity-level discrimination through lightweight contrastive training.
    
    \item Experimental results demonstrate WikiCLIP’s effectiveness, especially the strong generalization ability, achieving a state-of-the-art 28.5\% unseen accuracy on OVEN while reducing inference latency by nearly 100× compared with previous SOTA, AutoVER. 
\end{itemize}

\section{Related Works}

\subsection{Visual Entity Recognition}
Visual Entity Recognition (VER) aims to identify entities from visual inputs. Recent works have introduced large-scale datasets to advance open-domain entity recognition. OVEN~\citep{OVEN}, in particular, establishes a challenging benchmark with 6 million entity names, covering a diverse range of concepts.
Prior methods for open-domain VER can be broadly categorized into two paradigms based on their modeling approach: contrastive and generative.

\vspace{-2mm}
\paragraph{Contrastive-based Methods} Contrastive methods frame the problem as an image-to-text retrieval task, where dual-encoder models, such as CLIP-based architectures~\citep{CLIP,align}, are fine-tuned for specific entity recognition tasks. Recognition is based on selecting the top-scored result from the retrieved candidates. The OVEN~\citep{OVEN} dataset introduces two variants of CLIP, CLIPFusion and CLIP2CLIP, which are designed better to leverage multi-modal information from both queries and entities. Despite their effectiveness, these discriminative methods struggle to handle the semantic complexity and length discrepancies between encyclopedic descriptions and the simpler captions used during CLIP's pretraining~\citep{CLIP}. This limitation hinders the model's ability to adapt to and extract relevant information from structured descriptions. Concurrent work KnowCoL~\citep{zhou2025seeing} explores leveraging structural information by constructing a knowledge graph, which is orthogonal to our approach.

\begin{figure*}[ht]
    \centering
     \includegraphics[width=0.7\linewidth]{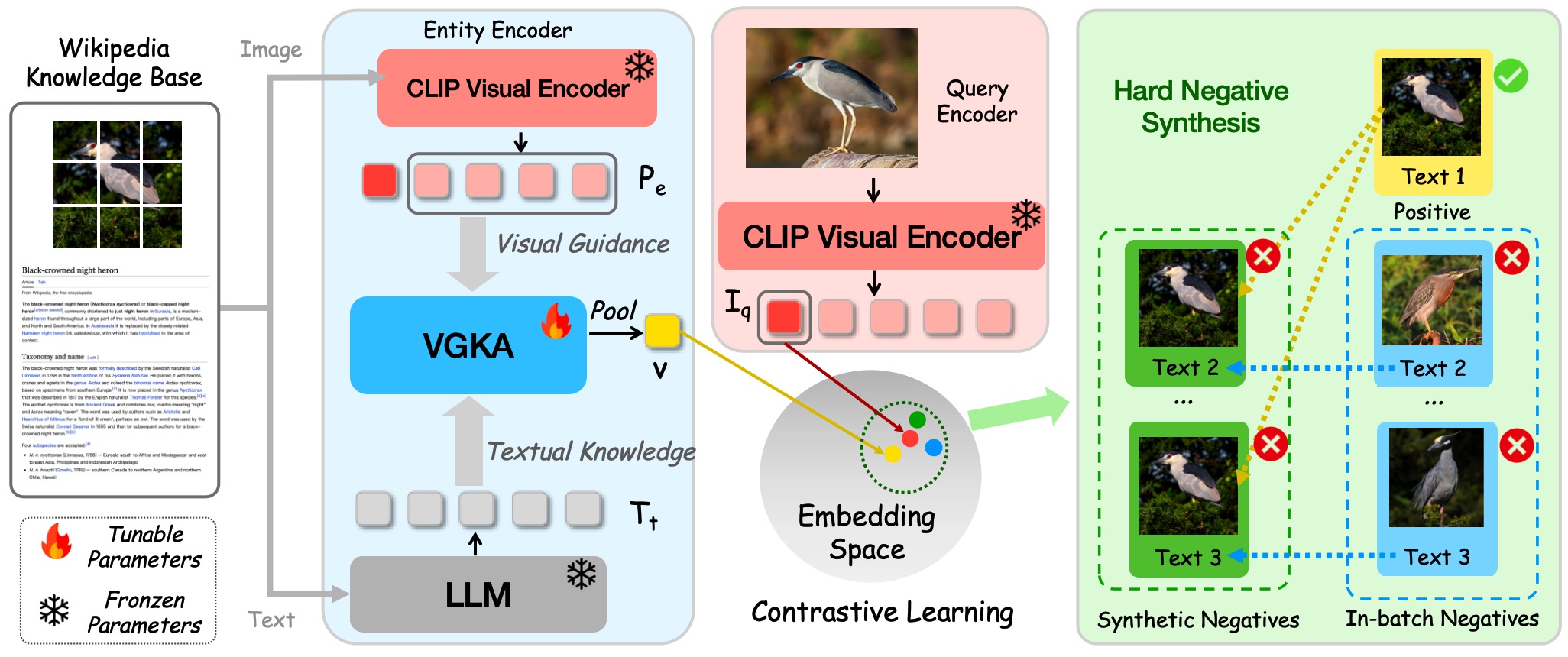}
    \caption{\textbf{The Overall Pipeline of WikiCLIP.} Given an entity's Wikipedia document, we use CLIP to extract patch-level features from the entity image and an LLM to obtain embeddings of its encyclopedic text description. The \textit{Vision-guided Knowledge Adaptation(VGKA)} selects the informative text tokens guided by the visual feature to produce an entity representation.
    To further improve fine-grained discrimination, we introduce a \textit{hard negative synthesis} strategy. This method generates challenging negative samples by replacing the original entity text description with that of a visually similar but semantically distinct entity. These synthetic hard negatives encourage the model to focus on subtle semantic differences.
    }
    \label{fig:pipeline}
    \vspace{-6mm}
\end{figure*}

\vspace{-2mm}
\paragraph{Generative-based Methods} In contrast to contrastive models, generative methods enable end-to-end visual entity recognition using auto-regressive language models. GER-ALD~\citep{GERALD} encodes Wikipedia entities into language-based discriminative (ALD) codes, which are then used as generation targets for visual entity prediction. Building on this idea, REW~\citep{REW} constructs large-scale entity caption datasets for web-scale VER, employing multimodal large language models for data verification and annotation.
AutoVER~\citep{autover} further advances the field by combining contrastive pretraining with a sequence-to-sequence generation objective, enabling fine-grained entity discrimination without requiring an external retriever. During inference, it restricts generation using retrieved candidates. While AutoVER adopts a visually grounded hard negative mining strategy, our approach focuses on synthesizing hard negatives via textual perturbations to enhance fine-grained semantic discrimination.
Despite their strong performance, generative models often suffer from high inference latency, limiting their practicality in real-world applications.
In this work, we address the limitations of both generative and contrastive methods by introducing a novel knowledge-aware entity representation and an efficient contrastive learning framework for visual entity recognition.

\vspace{-2mm}
\subsection{Cross-modal Representation Learning}
Multimodal representation learning has evolved mainly along two paradigms: contrastive and generative learning. Contrastive vision–language models such as CLIP~\citep{CLIP}, ALIGN~\citep{jia2021scaling}, and Florence~\citep{yuan2021florence} learn joint visual–textual embeddings by aligning paired image–text data on large-scale web corpora. Subsequent extensions, including FILIP~\citep{Yao2021FILIPFI}, VideoCLIP~\citep{xu2021videoclip}, and region-level alignment methods~\citep{zeng2021multi,kim2023region,maccap,HOICLIP,guo2022calip}, further explore multi-granular and temporal correspondence for enhanced cross-modal understanding.
In parallel, generative multimodal learning has emerged with models such as Flamingo~\citep{alayrac2022flamingo} and BLIP-2~\citep{li2023blip}, which integrate visual encoders with large language models to enable unified perception and reasoning. Follow-up frameworks like LLaVA~\citep{llava1.5,llava,liu2024llavanext,SPHINX,SPHINX-X} expand this paradigm through instruction tuning, demonstrating versatile multimodal interaction and understanding capabilities. More recently, reinforcement learning~\citep{dpo,grpo,noisygrpo,dadpo} has been used to improve the reasoning ability of MLLMs.
However, as these paradigms primarily emphasize global alignment and general reasoning, they struggle to encode the long-tailed, knowledge-intensive semantics required for fine-grained VER.
In this work, we introduce a lightweight framework that leverages the complementary strengths of contrastive and generative paradigms for better entity-level understanding.

\section{WikiCLIP}
In this section, we present WikiCLIP, a simple yet efficient framework for the challenging open-domain Visual Entity Recognition (VER), achieving strong performance with high inference efficiency. Our contributions are twofold: first, we design a \textit{Vision-Guided Knowledge Adaptor (VGKA)} that leverages LLM-based encyclopedic knowledge and selectively filters it with entity visual information, retaining entity-relevant content while discarding irrelevant information. Second, we propose a \textit{Hard Negative Synthesis} strategy, which enhances the discriminative power of the entity representations produced by VGKA during contrastive training, enabling more fine-grained entity recognition.
Specifically, we first introduce the task definition in Section~\ref{sec:definition}.
The model architecture is presented in Section~\ref{sec:overall}.  Then we introduce the contrastive learning with hard negative synthesis in Section~\ref{sec:training}. Finally, we introduce the VER inference pipeline in Section~\ref{sec:inf}.

\vspace{-2mm}
\subsection{Task definition}
\label{sec:definition}
We formulate Open-domain Visual Entity Recognition (VER) as a multimodal retrieval task. Given an image query \( I_q \), the goal is to retrieve the correct entity \( \mathbf{e} \) from a massive label space \( \mathcal{L} \) (e.g., \(\sim 6\)M entities in the OVEN~\citep{OVEN} Wikipedia KB). Each entity \( \mathbf{e} \in \mathcal{L} \) is associated with an entry in a knowledge base \( \mathcal{K} = \{(\mathbf{e}, \mathbf{E}_{\text{desc}}, \mathbf{E}_{\text{img}}) \mid \mathbf{e} \in \mathcal{L} \} \), where \( \mathbf{E}_{\text{desc}} \) is a textual description and \( \mathbf{E}_{\text{img}} = \{ I_e^1, I_e^2, \dots, I_e^{\mathbf{N}_e} \}\) is a set of $\mathbf{N}_e$ entity-associated images.



\subsection{Model Architecture}
\label{sec:overall}


WikiCLIP employs a dual-encoder architecture consisting of a frozen query image encoder and a trainable entity encoder, as illustrated in Figure~\ref{fig:pipeline}. The entity encoder incorporates a learnable Vision-Guided Knowledge Adaptor (VGKA), which integrates visual features from the frozen CLIP encoder with textual embeddings from the frozen LLM to produce knowledge-aware entity representations. The query image encoder is the standard CLIP visual encoder. The details are as follows:

\vspace{-2mm}
\paragraph{Vision-Guided Knowledge Adaptor}
In order to learn the knowledge-aware entity representations, we introduce a Vision-Guided Knowledge Adaptor which emphasizes encyclopedic text segments from the knowledge base that are semantically aligned with localized image regions, allowing the model to focus on entity-discriminative contents while filtering out irrelevant ones.

Specifically, we use the CLIP visual encoder to extract patch-level features \(\mathbf{P}_e \in \mathbb{R}^{N_p \times D}\) from the primary image \(I_e^1\) in \(\mathbf{E}_{img}\), serving as visual guidance, where \(N_p\) denotes the number of image patches and \(D\) is the dimension of the visual features.
Meanwhile, we employ a large language model (LLM) to encode the entity's associated text, yielding token-level embeddings. They are linearly projected to match the visual feature dimension via a learnable projection matrix \(\mathbf{W}_{\text{proj}} \in \mathbb{R}^{D_t \times D}\) to obtain the entity text representation \(\mathbf{T}_t \in \mathbb{R}^{N_t \times D}\), where \(D_t\) is the dimension of LLM embeddings and \(N_t\) is the length of text token.
The process is described as follows:
\begin{equation}
    \mathbf{P}_e = \mathrm{F}_{\text{CLIP}}(I_e),
\end{equation}
\begin{equation}
    \mathbf{T}_t = \mathrm{F}_{\text{LLM}}(\mathbf{E}_{desc}) \cdot \mathbf{W}_{\text{proj}},
\end{equation}
where \(\mathrm{F}_{\text{CLIP}}\) denotes the pretrained CLIP visual encoder and \(\mathrm{F}_{\text{LLM}}\) represents the large language model used for text encoding.

Given the extracted entity image and text representations, we then select discriminative text information from $\mathbf{T}_t$ with the guidance of visual features $\mathbf{P}_e$, using a multi-head cross attention operation. The output is the selected token-level entity representation $\mathbf{V'} \in \mathbb{R}^{N_t \times D}$. 
For clarity, we use $\mathrm{FA}(\mathrm{query}, \mathrm{key}, \mathrm{value}) = \mathrm{FFN}(\mathrm{MHA}(\mathrm{query}, \mathrm{key}, \mathrm{value}))$ to denote a multi-head attention block, where \(\mathrm{MHA}(\mathrm{query}, \mathrm{key}, \mathrm{value})\) is a multi-head attention, followed by a feed-forward network \(\mathrm{FFN}\). The process is described as follows:
\begin{equation}
    \mathbf{V'} = \mathrm{FA}(\mathbf{P}_e, \mathbf{T}_t, \mathbf{T}_t).
\end{equation}

To obtain an entity representation compatible with the CLIP embedding space, we apply mean pooling over the selected token-level entity representation $\mathbf{V'}$, resulting in a compact entity-level embedding $\mathbf{v} \in \mathbb{R}^D$. This aggregation preserves the overall semantics of the selected entity description as described below:
\begin{equation}
    \mathbf{v} = \mathrm{MeanPool}(\mathbf{V'}),
\end{equation}
where $\mathrm{MeanPool}$ denotes the average over token-level embeddings. In this way, our approach yields a knowledge-aware entity representation $\mathbf{v}$ by guiding the selection of textual information through visual cues. This representation integrates semantically rich textual knowledge with image-grounded attention, ensuring that the retained features are not only descriptive but also visually discriminative and entity-specific.

\subsection{Training with Hard Negative Synthesis}
\label{sec:training}
We employ a contrastive learning framework to optimize the knowledge-aware entity representations in WikiCLIP. First, we describe the overall contrastive learning pipeline tailored for the open-domain visual entity recognition task. Next, we present our hard negative synthesis strategy, designed to generate challenging negative samples that further enhance representation learning and discrimination.

\vspace{-2mm}
\paragraph{Contrastive Learning for VER}  
Our contrastive learning framework consists of three key steps:  
(1) \textit{Training Data Construction.} We construct training samples as multimodal triplets \((I_q, \mathbf{E}_{\text{img}}, \mathbf{E}_{\text{desc}})\), where \(I_q\) is the query image, and each entity is represented by its associated image \(\mathbf{E}_{\text{img}}\) and textual description \(\mathbf{E}_{\text{desc}}\).
(2) \textit{Query–Entity Encoding.} The query image \(I_q\) is encoded into a visual representation \(\mathbf{h}\) using the pretrained CLIP visual encoder. The entity \((\mathbf{E}_{\text{img}}, \mathbf{E}_{\text{desc}})\) is processed via the \textit{Vision-Guided Knowledge Adaptor}, which integrates both \(E_{\text{img}}\) and \(\mathbf{E}_{\text{desc}}\), producing the entity representation \(\mathbf{v}\).  
(3) \textit{Contrastive Learning Objective.} We optimize an InfoNCE~\citep{Oord2018RepresentationLW} loss to maximize similarity between matched query–entity pairs \((\mathbf{h}, \mathbf{v})\) while minimizing similarity with negative entity representations \(\mathbf{v}\). More details are provided in the supplementary material. 

Contrastive learning is typically performed at the batch level, where the hardness of negative samples within each batch significantly impacts training effectiveness. To enhance this, we propose a hard negative synthesis strategy, as described in the following section.

\vspace{-2mm}
\paragraph{Hard Negative Synthesis Strategy}
To improve fine-grained entity representation learning, we introduce a hard negative synthesis strategy. The key idea is to synthesize visually similar yet semantically mismatched negative samples, thereby strengthening the model’s sensitivity to textual nuances associated with similar visual appearances.

Specifically, we first leverage the CLIP visual features of the query image $H$ to construct minibatches where the query images within each batch are visually similar. For each sample in the visually clustered batches, we generate $N_{sync}$ synthetic entities composed of the original entity image and randomly selected textual descriptions from other entities in the minibatch. Then we obtain the representation of these synthetic entities for the $i^{th}$ query, denoted as $\mathcal{V}_i = \big\{ \mathbf{\tilde{v}}_i \big\}_{i=1}^{N_{\text{sync}}}$, where $\mathbf{\tilde{v}}_i \in \mathbb{R}^D$. These synthetic entity representations replace easy negative samples, those with low cosine similarity to the query image, thus enhancing the model's ability to learn more precise textual descriptions in entity representation learning. 

Formally, we define a mini-batch as
$\mathcal{B} = \big[(\mathbf{h}_i, \mathbf{v}_i)\big]_{i=1}^N$
where $\mathbf{h}^i$ is the query representation for the $i^{th}$ entity and $\mathbf{v}_i$ is the corresponding entity representation. Under the conventional contrastive learning paradigm, in-batch negatives for a query image $\mathbf{h}_i$ are defined as all entity representations $\mathbf{v}_j$ in the mini-batch $\mathcal{B}$ except for the positive pair. For the $i^{th}$ query, the in-batch negative $\mathcal{B}_i^-$ is defined as below.
\begin{equation}
\mathcal{B}^{-}_i = \{ \mathbf{v}_j \mid j \neq i, \ (\mathbf{h}_j, \mathbf{v}_j) \in \mathcal{B} \}.
\end{equation}

To increase the difficulty of negative samples, we \textit{selectively replace easy negatives with synthetic hard negatives}. Specifically, for each $\mathbf{v}_j \in \mathcal{B}^{-}_i$, we replace it with a synthetic negative entity feature $\mathbf{\tilde{v}}_j \in \mathcal{V}_i$ if:  
\begin{equation}
\mathrm{Sim}(\mathbf{h}_i, \mathbf{\tilde{v}}_j) > \mathrm{Sim}(\mathbf{h}_i, \mathbf{v}_j),
\end{equation}
where $\mathrm{Sim}(\cdot, \cdot)$ denotes cosine similarity. This replacement ensures that the synthetic hard negatives supplant the easier in-batch negatives, thereby encouraging the model to focus on more challenging examples and learn finer-grained discriminative features.

\subsection{Open-Domain VER Inference}

\label{sec:inf}
To perform visual entity recognition, we first compute entity representations for each entity in the knowledge base $(\mathbf{e}, \mathbf{E}_{\text{img}}, \mathbf{E}_{\text{desc}})$. For $I^i_e \in \mathbf{E}_{\text{img}}$, we extract an entity representation $\mathbf{v}^i$.
Given query image $I_q$, we use the CLIP visual encoder to extract the query representation $\mathbf{h}$. After that, the similarity score between $I_q$ and any entity $\mathbf{e}$ is computed as:
\begin{equation}
\mathrm{S}(I_q, \mathbf{e}) = \max_{i \in \{1, \dots, \mathbf{N}_e\}} \frac{\mathbf{h} \cdot \mathbf{v}^i}{\|\mathbf{h}\| \|\mathbf{v}^i\|}.
\end{equation}
After computing the similarity score between $\mathbf{h}$ and all entities $\mathbf{e}$, we choose the entity with the highest similarity score as the predicted entity.

Importantly, all entity embeddings in the knowledge base can be precomputed and stored. At inference time, recognition requires only a single similarity computation with the query image, unlike generative approaches that rely on autoregressive decoding, which is more computationally expensive.

\vspace{-2mm}
\section{Experiments}

This section presents a evaluation of our proposed method. We begin by detailing the experimental setup, then report and discuss the main results, followed by in-depth ablation studies and analysis to validate our design choices and probe the model's behavior.

\begin{table*}[]
\centering
\small
\begin{tabular}{cc|c|cc|ccc}
\toprule

 \textbf{Category}               &     \textbf{Methods}   & \textbf{Training}         & \multicolumn{2}{c|}{\textbf{Inference}} & \multicolumn{3}{c}{\textbf{OVEN}}      \\ \hline
                     &                 & Extra Dataset & Latency (ms)   & TFLOPS        & Unseen & Seen          & HM   \\ \hline
\multirow{2}{*}{\textit{Zero Shot}}
& GPT5-nano~\citep{openai2025gpt5}                 & -             & -              & -             & 13.0   & 23.7          & 16.8 \\ 
& GPT4V~\citep{openai2023gpt4v}                  & -             & -              & -             & \textbf{19.3}   & \textbf{29.8}          & \textbf{23.4} \\ 
\hline

\multirow{9}{*}{\textit{Generative}}  & PaLI-3B~\citep{pali}                & -             & -              & -             & 6.6    & 21.6          & 10.1 \\
                             & PaLI-17B~\citep{pali}               & -             & -              & -             & 12.4   & 30.6          & 17.6 \\
                             & $\text{GiT-Large}^{*}$~\citep{git} & WebLI-100M~\citep{pali}    & 83.95          & 3.06          & 4.2    & 13.7          & 6.5  \\
                             & $\text{GER-ALD}^{*}$~\citep{GERALD}   & Entity-WebLI~\citep{GERALD}  & 83.95          & 3.06          & 17.7   & 31.5          & 22.7 \\
                             & $\text{GiT-Large}^{*}$~\citep{git} & Entity-WebLI~\citep{GERALD}  & 83.95          & 3.06          & 16.4   & 25.9          & 20.1 \\
                             & $\text{GiT-Large}^{*}$~\citep{git} & REW-47M~\citep{REW}       & 83.95          & 3.06          & \textbf{25.1}   & 36            & 29.6 \\
                              & AutoVER 7B~\citep{autover}             & -             & 993            & 19.47         & 21.7   & 61.5          & 32.1 \\
                             & AutoVER 13B~\citep{autover}            & -             & 1569           & 24.74         & 24.5   & \textbf{63.6}          & \textbf{35.6} \\
                             \hline
\multirow{5}{*}{\textit{Contrastive}} & CLIP ViTL14~\citep{CLIP}            & -             & \textbf{11.69} & \textbf{0.07} & 5.4    & 5.3           & 5.4  \\
                             & CLIPFusion~\citep{OVEN}             & -             & 15.93          & 0.08          & 4.8    & 33.6          & 8.4  \\
                             & CLIP2CLIP~\citep{OVEN}              & -             & 13.84          & 0.08          & 10.5   & 12.6          & 11.5 \\
                             \rowcolor{blue!6}
                             & WikiCLIP-S             & -             & 14.49          & 1.93          & 27.0   & \textbf{36.8} & 31.1 \\
                             \rowcolor{blue!6}
                             & WikiCLIP-L             & -             & 14.49          & 1.93          & \textbf{28.5}   & 35.5          & \textbf{31.6} \\ \hline
\end{tabular}
    \caption{\textbf{Comparison with State-of-the-Art on the OVEN Entity Set.} 
    Methods categorized as \textit{zero-shot} correspond to general-purpose MLLMs that are not specifically trained for the VER task, while \textit{generative} and \textit{contrastive} paradigms denote models that are explicitly trained for Visual Entity Recognition. 
    Under the \textit{Train} columns, we list any additional training datasets beyond the OVEN training set (\textit{Extra Dataset}). For \textit{Inference}, we provide the inference \textit{latency}, measured on an A100 GPU, and report the \textit{TFLOPS}, which indicates the computational cost in floating-point operations. For baselines marked with $*$, results on the test split are reported due to the absence of validation split results.}
    \label{tab:main}
    \vspace{-4mm}
\end{table*}

\subsection{Training}
The WikiCLIP model is trained on a refined subset derived from the OVEN Entity training set. Following the established practice in~\citep{autover}, we construct a class-balanced subset by limiting each entity to at most 200 samples, resulting in 1M query-entity pairs spanning 7,943 entities. We further augment the training data with related multimodal Wikipedia documents in a self-supervised manner, bringing the total number of training samples to 1.9M. Implementation details regarding the selection and incorporation of Wikipedia documents are provided in the Supplementary.

\subsection{Evaluation Benchmarks}
We evaluate our methods on the Entity set of the OVEN~\citep{OVEN} benchmark, following prior works~\citep{autover,REW,GERALD}. 
The Entity set comprises 1,942 entities in the SEEN split and 1,182 in the UNSEEN split. We report top-1 VER prediction accuracy on each split, along with their harmonic mean (HM) as a combined performance measure.
The label space is derived from the English multimodal Wikipedia documents provided by E-VQA~\citep{EVQA}, encompassing 2 million entities.

To further assess the generalization ability of our method, we convert two downstream knowledge-intensive VQA datasets, namely E-VQA~\citep{EVQA} and INFOSEEK~\citep{infoseek}, into open-domain visual recognition tasks. 
In these tasks, a query image is paired with its corresponding Wikipedia document. We report accuracy on the SEEN, UNSEEN, and Overall splits for both datasets, where the split is defined by whether an entity appears in the training set.
For E-VQA, we use its official label space of 2 million multimodal Wikipedia documents. The test set comprises 2.1k entities, 500 of which are unseen. For INFOSEEK, we utilize a label space comprising 100k multimodal Wikipedia documents from Echosight~\citep{echosight}. The validation set contains 1.7k entities, 1.1k of which are unseen.

\subsection{Implementation Details}
For the visual encoder in WikiCLIP, we adopt the Vision Transformer (ViT) from EvaCLIP-8B~\citep{EVACLIP} to process images of size 224 × 224 pixels. The text encoder utilizes LLaMa3 series~\citep{llama3}. We set the maximum text token length $N_t$ to 256 and  truncate textual documents exceeding $N_t$ for efficiency. We provide two variants of WikiCLIP where \textit{WikiCLIP-S} is equipped with LLaMa3.2 1B and \textit{WikiCLIP-L} is equipped with LLaMa3.2 3B. A detailed analysis regarding model scale is provided in Section~\ref{sec:llm_scale}.

All of the parameters in the visual and text encoders are kept frozen during training. The only tunable \textit{Cross Attention} is a standard two-layer transformer~\citep{transformer} decoder with 0.08B parameters. We use a learning rate of 1e-4 with a cosine annealing decay strategy. The number of synthetic negative samples $N_{sync}$ is set to 8. 
The effective batch size is set to 128, and we train for a single epoch. The training process takes 19 hours using 8 A100 GPUs for WikiCLIP-S and 23 hours for WikiCLIP-L. We adopt the early stopping strategy based on the INFOSEEK results since only the Oven validation set is publicly available. During inference, we use the FAISS~\citep{faiss} library to match query and document features efficiently.



\begin{table}[]
\centering
\footnotesize
\begin{tabular}{cc|ccc}
\toprule
\textbf{Methods} & \textbf{Finetune Dataset} & \textbf{Unseen} & \textbf{Seen} & \textbf{Overall} \\  \midrule
\rowcolor{yellow!10} 
 \multicolumn{5}{c}{\textit{INFOSEEK}} \\  \midrule

DPR & In-house Data & - & - & 29.6 \\
$\text{CLIP I2T}^{*}$ & - & - & - & 32.0 \\
$\text{CLIP I2I}^{*}$ & - & 45.6 & 46.5 & 45.9 \\
Echosight & E-VQA & - & - & 53.2 \\
\rowcolor{blue!6}WikiCLIP-S & OVEN & 58.5 & 69.3 & 61.2 \\
\rowcolor{blue!6}WikiCLIP-L &  OVEN & \textbf{60.3} & \textbf{69.6} & \textbf{62.7} \\  \midrule
\rowcolor{yellow!10} 
 \multicolumn{5}{c}{\textit{E-VQA}} \\  \midrule
$\text{CLIP I2T}^{*}$ & - & - & - & 3.3 \\
$\text{CLIP I2I}^{*}$ & - & 14.6 & 10.6 & 13.3 \\
Echosight & E-VQA & - & - & 36.5 \\
Google Lens & Google Lens & - & - & \textbf{47.4} \\
\rowcolor{blue!6}WikiCLIP-S & OVEN & 27.7 & \textbf{39.9} & 30.7 \\
\rowcolor{blue!6}WikiCLIP-L & OVEN & \textbf{30.7} & 35.6 & 31.9 \\ \bottomrule
\end{tabular}
    \caption{\textbf{Performance on INFOSEEK and E-VQA.} The results marked with $*$ are provided by Echosight~\citep{echosight}.Google Lens is an image-to-image retrieval tool designed by Google.}
    \label{tab:vqa}
    \vspace{-4mm}
\end{table}


\subsection{Main Results}

\paragraph{Comparison with the State of the Arts on OVEN}
To demonstrate the effectiveness of WikiCLIP in large-scale visual entity recognition, we present evaluation results on the OVEN Entity validation set in Table~\ref{tab:main}. WikiCLIP shows a significant improvement across all metrics compared to contrastive-based methods. For example, it achieves 31.6 on HM, outperforming the previous contrastive SOTA, CLIP2CLIP~\citep{OVEN}, which scored 11.5. Additionally, WikiCLIP surpasses GiT-Large~\citep{git} trained on REW-47M~\citep{REW}, despite using only 1/5 of the tunable parameters and without relying on any other massive external datasets.
Moreover, WikiCLIP achieves superior generalization and much faster inference than large-scale alternatives. It significantly outperforms the 13B-parameter AutoVER~\citep{autover} with an unseen accuracy of 28.47 against 24.5, while reducing the inference latency by over two orders of magnitude to just 14.49 ms compared to 1569 ms.

\vspace{-3mm}
\paragraph{Generalization on E-VQA and INFOSEEK}
To further evaluate the generalization capability of WikiCLIP, we conduct experiments on two knowledge-intensive visual question answering benchmarks. As shown in Table~\ref{tab:vqa}, WikiCLIP achieves state-of-the-art performance on INFOSEEK~\citep{infoseek} without being fine-tuned on its training set. On E-VQA, WikiCLIP also delivers competitive results compared to Echosight~\citep{echosight}, despite the latter being explicitly fine-tuned on this dataset. These findings underscore the model's strong generalization capability across different datasets.


\vspace{-3mm}
\paragraph{Efficiency Comparison}
Despite the strong performance, our method demonstrates remarkable efficiency in both \textit{training cost} and \textit{inference speed}. As shown in Table~\ref{tab:main}, WikiCLIP has only 0.08B tunable parameters. Notably, unlike LoRA~\citep{hu2021lora}, our approach does not require gradient updates through the frozen LLM or CLIP, which reduces the training cost. The training time of our largest variant, WikiCLIP-L, is significantly lower than that of the previous SOTA method, AutoVER~\citep{autover} (13B), requiring only 23 hours compared to 247 hours on a node with 8 A100 GPUs. Furthermore, we train WikiCLIP using only the 1.9M training samples provided by OVEN~\citep{OVEN}, yet it outperforms the strong baseline GiT-Large, which is trained with REW~\citep{REW} on 47M carefully curated samples.  

Regarding inference speed, while WikiCLIP requires 6 hours on a node with 8 A100 GPUs to extract knowledge base features, the only computational cost at inference time is extracting the query feature and using FAISS to find the best match in the knowledge base. As a result, WikiCLIP achieves significantly lower inference latency than all generative-based methods—for example, \textit{14.49ms} vs. \textit{1569ms} when compared to AutoVER 13B, which relies on an auto-regressive generation process to predict a caption for a given query image.


\begin{table}[]
\setlength{\tabcolsep}{4pt}
\centering
\footnotesize
\begin{tabular}{cccc|ccc}
\toprule
\textbf{Image} & \textbf{Text} & \textbf{Cluster} & \textbf{Synthetic} & \textbf{Unseen} & \textbf{Seen} & \textbf{Overall} \\ \midrule
\rowcolor{yellow!10}
 \multicolumn{7}{c}{\textit{Entity Representation}} \\ \hline
\checkmark &  &  &  & 39.5 & 60.4 & 44.8 \\
 & \checkmark &  &  & 47.9 & 59.1 & 50.8 \\
\checkmark & \checkmark &  &  & 56.8 & 68.0 & 59.7 \\ \midrule
\rowcolor{yellow!10}
 \multicolumn{7}{c}{\textit{Training Strategy}} \\ \hline
\checkmark & \checkmark & \checkmark &  & 56.8 & 68.2 & 59.7 \\
\checkmark & \checkmark &  & \checkmark & 57.0 & 64.6 & 58.9 \\
\rowcolor{blue!6}\checkmark & \checkmark & \checkmark & \checkmark & \textbf{58.5} & \textbf{69.3} & \textbf{61.2} \\ \bottomrule
\end{tabular}
\caption{\textbf{Ablation Study on Entity Representation and Training Strategy.} The \textit{Image} and \textit{Text} refer to the use of images and text from Wikipedia documents for extracting visual features. The \textit{Cluster} and \textit{Synthetic} represent the visual-based clustering and synthetic generation of hard textual negatives within the proposed hard negative synthesis.}
    \vspace{-5mm}

\label{tab:ablation}
\end{table}

\subsection{Ablation Study}
\vspace{-1mm}
To provide a comprehensive understanding of WikiCLIP, we conduct an ablation study to show the effect of WikiCLIP's architecture and training strategy. 
Our ablation utilizes the INFOSEEK dataset~\citep{infoseek} with a 100k knowledge base. 
More ablation experiment results on OVEN and EVQA are provided in the Supplementary.

\vspace{-4mm}
\paragraph{Entity Representation}
We perform an ablation study on the input modalities of our Entity Encoder, evaluating three configurations: using only document images, only textual descriptions, and our full model that leverages both. As shown in Table~\ref{tab:ablation}, we observe that excluding detailed textual descriptions results in entity representations that lack discriminative power for fine-grained categories, leading to performance degradation. Conversely, relying solely on text creates a significant representational gap between the query image and the entity, which also impairs performance.

\vspace{-3mm}
\paragraph{Training Strategy}
The \textit{Hard Negative Synthesis Strategy} consists of two sequential steps: visual-based clustering and synthetic hard negative sample generation. As shown in Table~\ref{tab:ablation}, applying either step individually does not yield a significant performance improvement. Only when both steps are combined can we construct an effective hard negative batch. We conclude that visual-based clustering alone groups visually similar samples within a batch but fails to provide sufficiently fine-grained hard negatives for open-domain visual entity recognition. By building upon these visually similar samples, the synthetic hard negative sample generation introduces fine-grained variations, where the synthetic entities retain the same image but incorporate different textual descriptions, offering more effective guidance. We provide visualization results of entity representation in the Supplementary.

\begin{table}[]
\setlength{\tabcolsep}{4pt}
\centering
\small
\begin{tabular}{ccccc}
\toprule
\textbf{Text Encoder} & \textbf{Visual Encoder} & \textbf{Unseen} & \textbf{Seen} & \textbf{Overall } \\ \hline
EVA-CLIP 8B & CLIP ViTL & 26.5 & 54.6 & 33.6 \\
LLaMa3.2 1B & CLIP ViTL & 39.8 & 46.9 & 41.6 \\ 
EVA-CLIP 8B & EVA-CLIP 8B & 56.3 & 62.4 & 58.1 \\
\rowcolor{blue!6}LLaMa3.2 1B & EVA-CLIP 8B & \textbf{58.5} & \textbf{69.3} & \textbf{61.2} \\
\hline
\end{tabular}
\caption{\textbf{Performance comparison with different text and visual encoders.} We report the accuracy on the INFOSEEK validation set. The \textit{CLIP ViTL} is provided by OpenAI~\citep{CLIP} and contains 0.3B parameters.}
\vspace{-4mm}
\label{tab:encoders}
\vspace{-2mm}
\end{table}

\begin{figure*}[ht]
    \centering
     \includegraphics[width=0.55\linewidth]{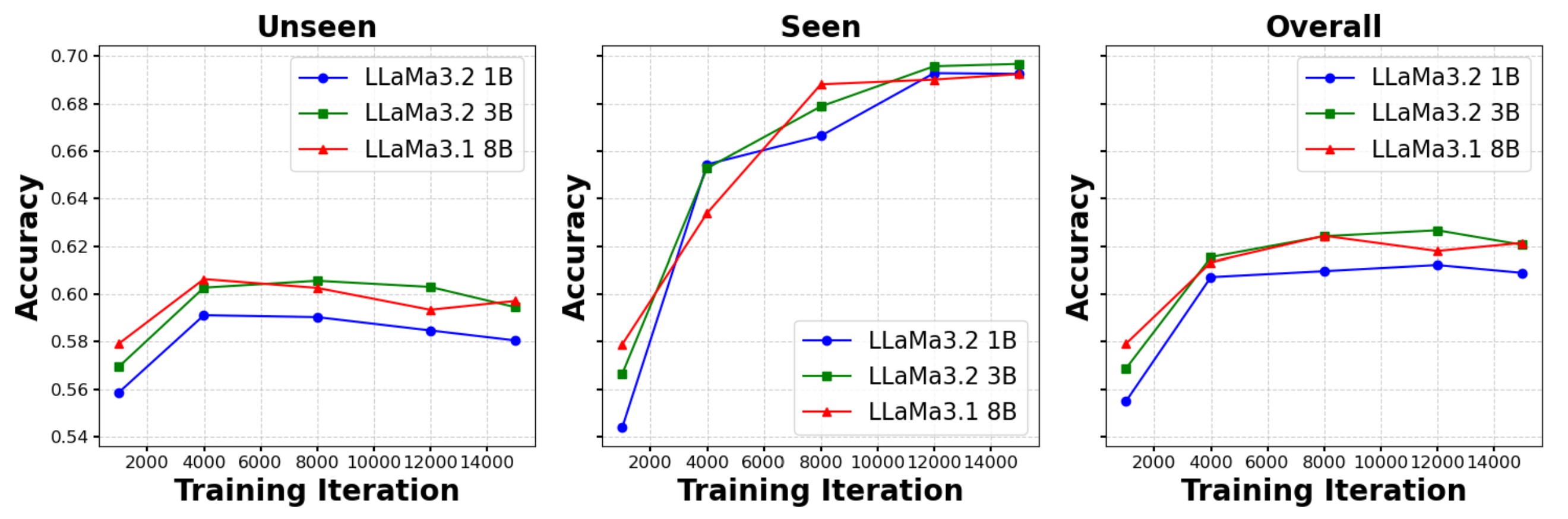}   
     \vspace{-4mm}
    \caption{\textbf{Performance with varying training iterations and LLM choices.} We report the accuracy of the INFOSEEK validation set of WikiCLIP using three different scales of LLMs, along with varying training iterations.
    }
    \label{fig:scale}
    \vspace{-4mm}
\end{figure*}


\vspace{-4mm}
\paragraph{Choice of Visual and Textual Encoders}
We conduct an ablation study on the choice of visual and textual encoders to examine their impact on WikiCLIP. For the textual encoder, we compare a large language model with the CLIP text encoder. For the visual encoder, we evaluate CLIP variants with varying parameter scales. As shown in Table~\ref{tab:encoders}, the visual encoder plays a crucial role in WikiCLIP, as the frozen visual embedding space serves as the matching space for entity representation. Additionally, the CLIP text encoder proves to be suboptimal compared to the LLM, which incorporates richer world knowledge and can process longer text sequences beyond CLIP’s token limit.


\subsection{Analysis and Discussion}
\label{sec4.6}

\begin{figure}[t]
    \centering
     \includegraphics[width=0.6\linewidth]{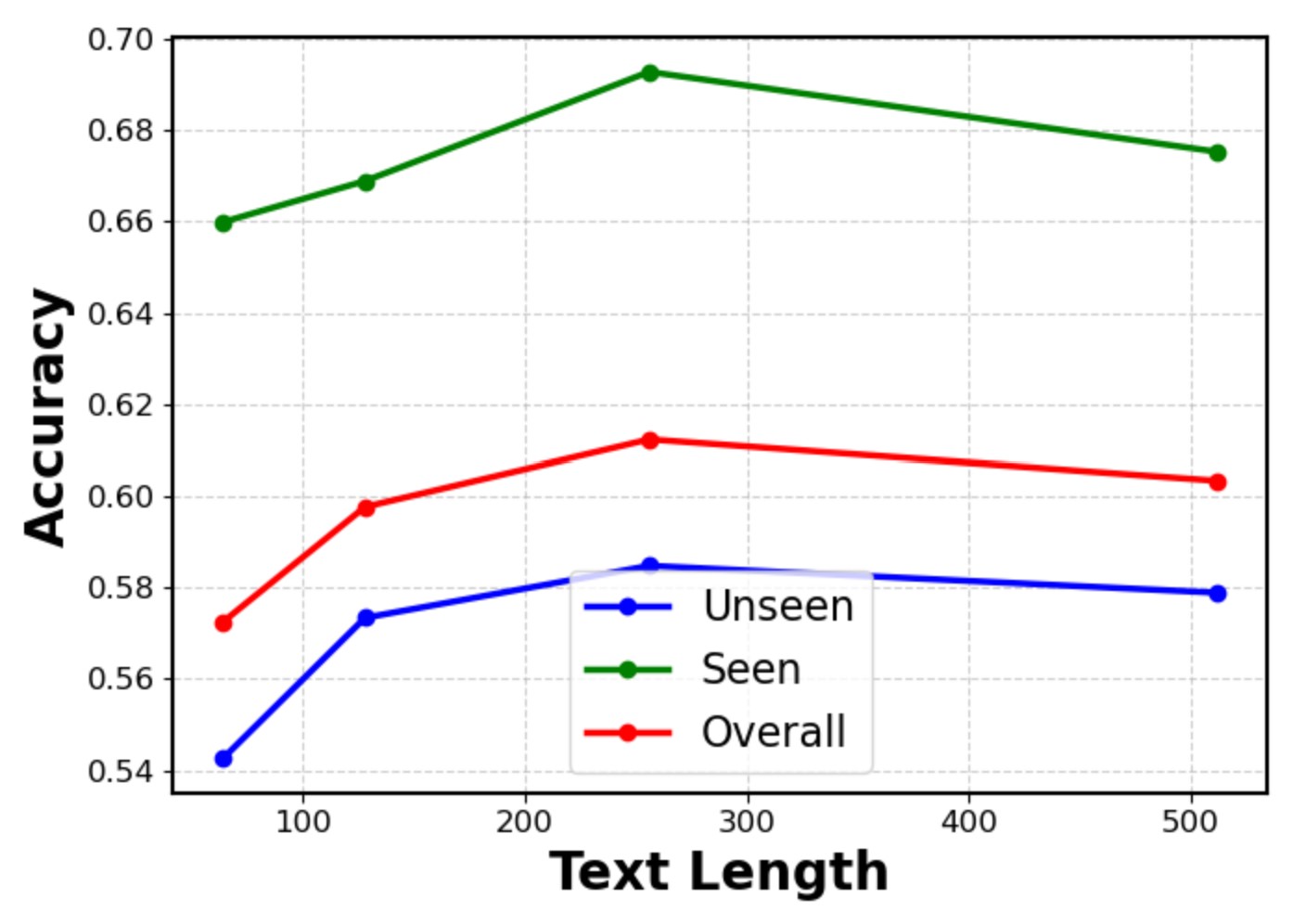}
     \vspace{-4mm}
    \caption{\textbf{Performance with different text length.} We report the accuracy of WikiCLIP-S on the INFOSEEK validation set. The best performance is achieved at 256 text length.
    }
    \label{fig:length}
    \vspace{-4mm}
\end{figure}

In this section, we provide empirical analysis to explore the key properties of WikiCLIP and provide insights for further research on contrastive-based visual entity recognition.

\vspace{-4mm}
\paragraph{Does the Length of Encoded Wiki Text Matter?}
One of the key contributions of WikiCLIP is its ability to extract structured textual knowledge that differentiates similar entity categories, raising the question of how much text ultimately contributes to performance. To investigate this, we conduct experiments to examine the impact of encoded Wikipedia text length on model performance. As shown in Figure~\ref{fig:length}, performance improves as text length increases, peaking at a length of 256 tokens. This phenomenon suggests that not all textual information in Wikipedia documents is beneficial for visual recognition; excessive text introduces noise, which hampers the extraction of structured knowledge. Furthermore, this suggests a potential avenue for enhancing WikiCLIP by focusing on extracting more informative text from Wikipedia documents, which we leave for future research.

\vspace{-4mm}
\paragraph{Does the Stronger LLM Improve the Performance?}
\label{sec:llm_scale}
Beyond text length, another critical aspect of structured knowledge extraction is whether a more powerful LLM enhances performance. To investigate this, we conduct experiments using three LLMs of different scales: \textit{LLaMa-3.2 1B}, \textit{LLaMa-3.2 3B}, and \textit{LLaMa-3.1 8B}. As shown in Figure~\ref{fig:scale} in the Supplementary, we observe that increasing model size generally improves performance, particularly in unseen accuracy, with the 1B LLM consistently underperforming compared to the 3B and 8B variants. However, the performance gains between the 3B and 8B models are marginal. This suggests that scaling up LLMs primarily enhances generative capabilities, but does not necessarily improve representation quality for visual recognition. A promising direction for future work is to develop a dedicated text embedding model optimized for extracting encyclopedic knowledge.

\vspace{-4mm}

\begin{figure}[t]
    \centering
     \includegraphics[width=0.55\linewidth]{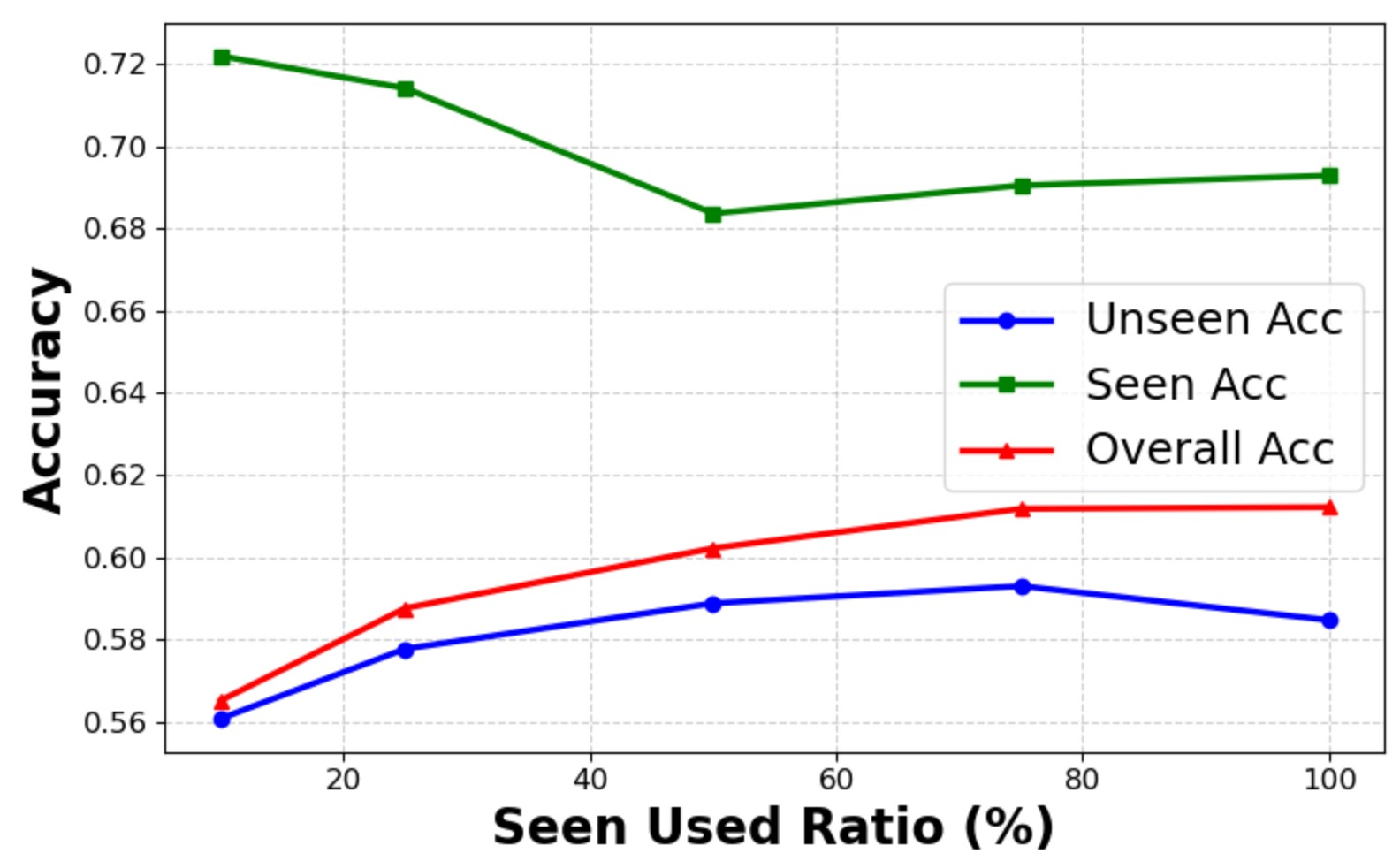}   
          \vspace{-3mm}
    \caption{\textbf{Performance with Different Ratios of Seen Entities.} We evaluate models trained with varying ratios of seen entities. \textit{Seen Acc} and \textit{Unseen Acc} measure accuracy on test samples whose entities were present or absent, respectively, during training.}
    \label{fig:seen_ratio}
    \vspace{-5mm}
\end{figure}

\paragraph{Influence of the Number of Seen Categories.}
To further examine the generalization ability of WikiCLIP, we conduct experiments using different numbers of seen entity training samples. The OVEN~\citep{OVEN} entity training set consists of 7,943 entities. We create new training sets by sampling \textit{10\%}, \textit{25\%}, \textit{50\%}, and \textit{75\%} of the original seen entities. As shown in Figure~\ref{fig:seen_ratio}, WikiCLIP demonstrates strong generalization capability even with only 700 seen entities, achieving an unseen accuracy of 56\%, compared to 58\% when trained with the full 7,943 seen entities. This characteristic is particularly valuable for open-domain visual entity recognition, where collecting sufficient training samples to cover the 2M entities in Wikipedia is infeasible.

\vspace{-2mm}
\section{Conclusion}
\vspace{-1mm}
In this work, we present WikiCLIP, a simple yet efficient framework for open-domain visual entity recognition. 
WikiCLIP employs a Vision-Guided Knowledge Adaptor to extract discriminative entity representation and a hard negative synthesis strategy to generate challenging negatives for the contrastive training of VGKA. 
Extensive experiments on standard open-domain VER benchmarks show that WikiCLIP substantially outperforms strong baselines, achieving significant performance gains while maintaining fast inference speed.

\vspace{-4mm}
\paragraph{Limitation} While WikiCLIP provides a simple and efficient baseline for open-domain VER, it still underutilizes the knowledge encoded in LLMs. Our analysis shows that performance quickly saturates with longer text inputs and that scaling to larger LLMs yields only marginal gains. These findings suggest that expanding model size or context length alone cannot fully exploit LLM-guided contrastive learning, which motivate future work on better leveraging LLM knowledge and refining entity representations.

\section{Acknowledgments}
This work was supported by NSFC 62350610269, Shanghai Frontiers Science Center of Human-centered Artificial Intelligence, and MoE Key Lab of Intelligent Perception and Human-Machine Collaboration (ShanghaiTech University). This work was also supported by HPC platform of ShanghaiTech University.
{
    \small
    \bibliographystyle{ieeenat_fullname}
    \bibliography{main}

@String(CVPR= {IEEE Conf. Comput. Vis. Pattern Recog.})

@String(ICCV= {Int. Conf. Comput. Vis.})

@String(CVPR  = {CVPR})

@String(ICCV  = {ICCV})

@article{alayrac2022flamingo,
  title={Flamingo: a visual language model for few-shot learning},
  author={Alayrac, Jean-Baptiste and Donahue, Jeff and Luc, Pauline and Miech, Antoine and Barr, Iain and Hasson, Yana and Lenc, Karel and Mensch, Arthur and Millican, Katherine and Reynolds, Malcolm and others},
  journal={Advances in Neural Information Processing Systems},
  volume={35},
  pages={23716--23736},
  year={2022}
}

@article{li2023blip,
  title={Blip-2: Bootstrapping language-image pre-training with frozen image encoders and large language models},
  author={Li, Junnan and Li, Dongxu and Savarese, Silvio and Hoi, Steven},
  journal={arXiv preprint arXiv:2301.12597},
  year={2023}
}

@online{openai2025gpt5,
  author    = {OpenAI},
  title     = {GPT-5: Advancing General-Purpose Language Intelligence},
  year      = {2025},
  url       = {https://openai.com/research/gpt-5},
  urldate   = {2025-11-12},
  note      = {Accessed on November 12, 2025}
}

@article{hu2021lora,
  title={Lora: Low-rank adaptation of large language models},
  author={Hu, Edward J and Shen, Yelong and Wallis, Phillip and Allen-Zhu, Zeyuan and Li, Yuanzhi and Wang, Shean and Wang, Lu and Chen, Weizhu},
  journal={arXiv preprint arXiv:2106.09685},
  year={2021}
}

@article{guo2022calip,
  title={Calip: Zero-shot enhancement of clip with parameter-free attention},
  author={Guo, Ziyu and Zhang, Renrui and Qiu, Longtian and Ma, Xianzheng and Miao, Xupeng and He, Xuming and Cui, Bin},
  journal={arXiv preprint arXiv:2209.14169},
  year={2022}
}

@inproceedings{transformer,
  title={Attention is all you need},
  author={Vaswani, Ashish and Shazeer, Noam and Parmar, Niki and Uszkoreit, Jakob and Jones, Llion and Gomez, Aidan N and Kaiser, {\L}ukasz and Polosukhin, Illia},
  booktitle={Advances in neural information processing systems},
  pages={5998--6008},
  year={2017}
}

@misc{openai2023gpt4v,
  author={OpenAI},
    title = {Vision - OpenAI API},
    howpublished = {\url{https://platform.openai.com/docs/guides/vision}},
    year = {2023},
}

@article{llava,
  title={Visual instruction tuning},
  author={Liu, Haotian and Li, Chunyuan and Wu, Qingyang and Lee, Yong Jae},
  journal={arXiv preprint arXiv:2304.08485},
  year={2023}
}

@inproceedings{jia2021scaling,
  title={Scaling up visual and vision-language representation learning with noisy text supervision},
  author={Jia, Chao and Yang, Yinfei and Xia, Ye and Chen, Yi-Ting and Parekh, Zarana and Pham, Hieu and Le, Quoc and Sung, Yun-Hsuan and Li, Zhen and Duerig, Tom},
  booktitle={International conference on machine learning},
  pages={4904--4916},
  year={2021},
  organization={PMLR}
}

@inproceedings{CLIP,
  title={Learning transferable visual models from natural language supervision},
  author={Radford, Alec and Kim, Jong Wook and Hallacy, Chris and Ramesh, Aditya and Goh, Gabriel and Agarwal, Sandhini and Sastry, Girish and Askell, Amanda and Mishkin, Pamela and Clark, Jack and others},
  booktitle={International conference on machine learning},
  pages={8748--8763},
  year={2021},
  organization={PMLR}
}

@article{yuan2021florence,
  title={Florence: A new foundation model for computer vision},
  author={Yuan, Lu and Chen, Dongdong and Chen, Yi-Ling and Codella, Noel and Dai, Xiyang and Gao, Jianfeng and Hu, Houdong and Huang, Xuedong and Li, Boxin and Li, Chunyuan and others},
  journal={arXiv preprint arXiv:2111.11432},
  year={2021}
}

@article{xu2021videoclip,
  title={Videoclip: Contrastive pre-training for zero-shot video-text understanding},
  author={Xu, Hu and Ghosh, Gargi and Huang, Po-Yao and Okhonko, Dmytro and Aghajanyan, Armen and Metze, Florian and Zettlemoyer, Luke and Feichtenhofer, Christoph},
  journal={arXiv preprint arXiv:2109.14084},
  year={2021}
}

@article{llava1.5,
  title={Improved Baselines with Visual Instruction Tuning},
  author={Haotian Liu and Chunyuan Li and Yuheng Li and Yong Jae Lee},
  journal={ArXiv},
  year={2023},
  volume={abs/2310.03744},
}

@inproceedings{SPHINX,
  title={Sphinx: A mixer of weights, visual embeddings and image scales for multi-modal large language models},
  author={Lin, Ziyi and Liu, Dongyang and Zhang, Renrui and Gao, Peng and Qiu, Longtian and Xiao, Han and Qiu, Han and Shao, Wenqi and Chen, Keqin and Han, Jiaming and others},
  booktitle={European Conference on Computer Vision},
  pages={36--55},
  year={2024},
  organization={Springer}
}

@article{SPHINX-X,
  title={SPHINX-X: Scaling Data and Parameters for a Family of Multi-modal Large Language Models},
  author={Peng Gao and Renrui Zhang and Chris Liu and Longtian Qiu and Siyuan Huang and Weifeng Lin and Shitian Zhao and Shijie Geng and Ziyi Lin and Peng Jin and Kaipeng Zhang and Wenqi Shao and Chao Xu and Conghui He and Junjun He and Hao Shao and Pan Lu and Hongsheng Li and Yu Qiao},
  journal={ArXiv},
  year={2024},
  volume={abs/2402.05935},
}

@article{align,
  title={Scaling Up Visual and Vision-Language Representation Learning With Noisy Text Supervision},
  author={Chao Jia and Yinfei Yang and Ye Xia and Yi-Ting Chen and Zarana Parekh and Hieu Pham and Quoc V. Le and Yun-Hsuan Sung and Zhen Li and Tom Duerig},
  journal={ArXiv},
  year={2021},
  volume={abs/2102.05918},
}

@article{EVQA,
  title={Encyclopedic VQA: Visual questions about detailed properties of fine-grained categories},
  author={Thomas Mensink and Jasper R. R. Uijlings and Llu{\'i}s Castrej{\'o}n and Arushi Goel and Felipe Cadar and Howard Zhou and Fei Sha and Andre F. de Ara{\'u}jo and Vittorio Ferrari},
  journal={2023 IEEE/CVF International Conference on Computer Vision (ICCV)},
  year={2023},
  pages={3090-3101},
}

@article{infoseek,
  title={Can Pre-trained Vision and Language Models Answer Visual Information-Seeking Questions?},
  author={Yang Chen and Hexiang Hu and Yi Luan and Haitian Sun and Soravit Changpinyo and Alan Ritter and Ming-Wei Chang},
  journal={ArXiv},
  year={2023},
  volume={abs/2302.11713},
}

@article{echosight,
  title={EchoSight: Advancing Visual-Language Models with Wiki Knowledge},
  author={Yibin Yan and Weidi Xie},
  journal={ArXiv},
  year={2024},
  volume={abs/2407.12735},
}

@article{llama3,
  title={The Llama 3 Herd of Models},
  author={Abhimanyu Dubey and Abhinav Jauhri and et al.},
  journal={ArXiv},
  year={2024},
  volume={abs/2407.21783},
}

@inproceedings{autover,
  title={Grounding language models for visual entity recognition},
  author={Xiao, Zilin and Gong, Ming and Cascante-Bonilla, Paola and Zhang, Xingyao and Wu, Jie and Ordonez, Vicente},
  booktitle={European Conference on Computer Vision},
  pages={393--411},
  year={2024},
  organization={Springer}
}

@article{REW,
  title={Web-Scale Visual Entity Recognition: An LLM-Driven Data Approach},
  author={Mathilde Caron and Alireza Fathi and Cordelia Schmid and Ahmet Iscen},
  journal={ArXiv},
  year={2024},
  volume={abs/2410.23676},
}

@article{GERALD,
  title={A Generative Approach for Wikipedia-Scale Visual Entity Recognition},
  author={Mathilde Caron and Ahmet Iscen and Alireza Fathi and Cordelia Schmid},
  journal={2024 IEEE/CVF Conference on Computer Vision and Pattern Recognition (CVPR)},
  year={2024},
  pages={17313-17322},
}

@article{OVEN,
  title={Open-domain Visual Entity Recognition: Towards Recognizing Millions of Wikipedia Entities},
  author={Hexiang Hu and Yi Luan and Yang Chen and Urvashi Khandelwal and Mandar Joshi and Kenton Lee and Kristina Toutanova and Ming-Wei Chang},
  journal={2023 IEEE/CVF International Conference on Computer Vision (ICCV)},
  year={2023},
  pages={12031-12041},
}

@article{EVACLIP,
  title={EVA-CLIP: Improved Training Techniques for CLIP at Scale},
  author={Quan Sun and Yuxin Fang and Ledell Yu Wu and Xinlong Wang and Yue Cao},
  journal={ArXiv},
  year={2023},
  volume={abs/2303.15389},
}

@article{faiss,
      title={The Faiss library},
      author={Matthijs Douze and Alexandr Guzhva and Chengqi Deng and Jeff Johnson and Gergely Szilvasy and Pierre-Emmanuel Mazaré and Maria Lomeli and Lucas Hosseini and Hervé Jégou},
      year={2024},
      eprint={2401.08281},
      archivePrefix={arXiv},
      primaryClass={cs.LG}
}

@article{git,
  title={GIT: A Generative Image-to-text Transformer for Vision and Language},
  author={Jianfeng Wang and Zhengyuan Yang and Xiaowei Hu and Linjie Li and Kevin Lin and Zhe Gan and Zicheng Liu and Ce Liu and Lijuan Wang},
  journal={ArXiv},
  year={2022},
  volume={abs/2205.14100},
}

@article{pali,
  title={PaLI: A Jointly-Scaled Multilingual Language-Image Model},
  author={Xi Chen and Xiao Wang and Soravit Changpinyo and A. J. Piergiovanni and Piotr Padlewski and Daniel M. Salz and Sebastian Goodman and Adam Grycner and Basil Mustafa and Lucas Beyer and Alexander Kolesnikov and Joan Puigcerver and Nan Ding and Keran Rong and Hassan Akbari and Gaurav Mishra and Linting Xue and Ashish V. Thapliyal and James Bradbury and Weicheng Kuo and Mojtaba Seyedhosseini and Chao Jia and Burcu Karagol Ayan and Carlos Riquelme and Andreas Steiner and Anelia Angelova and Xiaohua Zhai and Neil Houlsby and Radu Soricut},
  journal={ArXiv},
  year={2022},
  volume={abs/2209.06794},
}

@inproceedings{Wah2011TheCB,
  title={The Caltech-UCSD Birds-200-2011 Dataset},
  author={Catherine Wah and Steve Branson and Peter Welinder and Pietro Perona and Serge J. Belongie},
  year={2011},
}

@article{Horn2017TheIS,
  title={The iNaturalist Species Classification and Detection Dataset},
  author={Grant Van Horn and Oisin Mac Aodha and Yang Song and Yin Cui and Chen Sun and Alexander Shepard and Hartwig Adam and Pietro Perona and Serge J. Belongie},
  journal={2018 IEEE/CVF Conference on Computer Vision and Pattern Recognition},
  year={2017},
  pages={8769-8778},
}

@inproceedings{Khosla2012NovelDF,
  title={Novel Dataset for Fine-Grained Image Categorization : Stanford Dogs},
  author={Aditya Khosla and Nityananda Jayadevaprakash and Bangpeng Yao and Li Fei-Fei},
  year={2012},
}

@inproceedings{Biten_Gomez_Rusinol_Karatzas_2019,   title={Good News, Everyone! Context driven entity-aware captioning for news images},  url={http://dx.doi.org/10.1109/cvpr.2019.01275},  DOI={10.1109/cvpr.2019.01275},  booktitle={2019 IEEE/CVF Conference on Computer Vision and Pattern Recognition (CVPR)},  author={Biten, Ali Furkan and Gomez, Lluis and Rusinol, Marcal and Karatzas, Dimosthenis},  year={2019},  month={Jun},  language={en-US}  }

@inproceedings{Fu_Wang_Yang_2021,   title={MM-AVS: A Full-Scale Dataset for Multi-modal Summarization},  url={http://dx.doi.org/10.18653/v1/2021.naacl-main.473},  DOI={10.18653/v1/2021.naacl-main.473},  booktitle={Proceedings of the 2021 Conference of the North American Chapter of the Association for Computational Linguistics: Human Language Technologies},  author={Fu, Xiyan and Wang, Jun and Yang, Zhenglu},  year={2021},  month={Jan},  language={en-US}  }

@inproceedings{Liu_Wang_Wang_Ordonez_2021,   title={Visual News: Benchmark and Challenges in News Image Captioning},  url={http://dx.doi.org/10.18653/v1/2021.emnlp-main.542},  DOI={10.18653/v1/2021.emnlp-main.542},  booktitle={Proceedings of the 2021 Conference on Empirical Methods in Natural Language Processing},  author={Liu, Fuxiao and Wang, Yinghan and Wang, Tianlu and Ordonez, Vicente},  year={2021},  month={Jan},  language={en-US}  }

@article{Oord2018RepresentationLW,
  title={Representation Learning with Contrastive Predictive Coding},
  author={A{\"a}ron van den Oord and Yazhe Li and Oriol Vinyals},
  journal={ArXiv},
  year={2018},
  volume={abs/1807.03748},
  url={https://api.semanticscholar.org/CorpusID:49670925}
}

@inproceedings{kim2023region,
  title={Region-aware pretraining for open-vocabulary object detection with vision transformers},
  author={Kim, Dahun and Angelova, Anelia and Kuo, Weicheng},
  booktitle={Proceedings of the IEEE/CVF conference on computer vision and pattern recognition},
  pages={11144--11154},
  year={2023}
}

@article{zeng2021multi,
  title={Multi-grained vision language pre-training: Aligning texts with visual concepts},
  author={Zeng, Yan and Zhang, Xinsong and Li, Hang},
  journal={arXiv preprint arXiv:2111.08276},
  year={2021}
}

@article{Yao2021FILIPFI,
  title={FILIP: Fine-grained Interactive Language-Image Pre-Training},
  author={Lewei Yao and Runhu Huang and Lu Hou and Guansong Lu and Minzhe Niu and Hang Xu and Xiaodan Liang and Zhenguo Li and Xin Jiang and Chunjing Xu},
  journal={ArXiv},
  year={2021},
  volume={abs/2111.07783},
}

@misc{liu2024llavanext,
    title={LLaVA-NeXT: Improved reasoning, OCR, and world knowledge},
    url={https://llava-vl.github.io/blog/2024-01-30-llava-next/},
    author={Liu, Haotian and Li, Chunyuan and Li, Yuheng and Li, Bo and Zhang, Yuanhan and Shen, Sheng and Lee, Yong Jae},
    month={January},
    year={2024}
}

@article{dadpo,
  title={DA-DPO: Cost-efficient Difficulty-aware Preference Optimization for Reducing MLLM Hallucinations},
  author={Qiu, Longtian and Ning, Shan and Zhang, Chuyu and Sun, Jiaxuan and He, Xuming},
  journal={arXiv preprint arXiv:2601.00623},
  year={2026}
}

@article{ning2026wiki,
  title={Wiki-R1: Incentivizing Multimodal Reasoning for Knowledge-based VQA via Data and Sampling Curriculum},
  author={Ning, Shan and Qiu, Longtian and He, Xuming},
  journal={arXiv preprint arXiv:2603.05256},
  year={2026}
}

@article{maccap,
  title={Mining Fine-Grained Image-Text Alignment for Zero-Shot Captioning via Text-Only Training},
  author={Longtian Qiu and Shan Ning and Xuming He},
  journal={ArXiv},
  year={2024},
  volume={abs/2401.02347},
}

@article{HOICLIP,
  title={HOICLIP: Efficient Knowledge Transfer for HOI Detection with Vision-Language Models},
  author={Sha Ning and Longtian Qiu and Yongfei Liu and Xuming He},
  journal={2023 IEEE/CVF Conference on Computer Vision and Pattern Recognition (CVPR)},
  year={2023},
  pages={23507-23517},
}

@article{noisygrpo,
  title={NoisyGRPO: Incentivizing Multimodal CoT Reasoning via Noise Injection and Bayesian Estimation},
  author={Qiu, Longtian and Ning, Shan and Sun, Jiaxuan and He, Xuming},
  journal={Advances in Neural Information Processing Systems},
  volume={38},
  pages={124239--124267},
  year={2026}
}

@article{grpo,
  title={Deepseekmath: Pushing the limits of mathematical reasoning in open language models},
  author={Shao, Zhihong and Wang, Peiyi and Zhu, Qihao and Xu, Runxin and Song, Junxiao and Bi, Xiao and Zhang, Haowei and Zhang, Mingchuan and Li, YK and Wu, Yang and others},
  journal={arXiv preprint arXiv:2402.03300},
  year={2024}
}

@article{dpo,
  title={Direct preference optimization: Your language model is secretly a reward model},
  author={Rafailov, Rafael and Sharma, Archit and Mitchell, Eric and Manning, Christopher D and Ermon, Stefano and Finn, Chelsea},
  journal={Advances in neural information processing systems},
  volume={36},
  pages={53728--53741},
  year={2023}
}

@article{zhou2025seeing,
  title={Seeing and Knowing in the Wild: Open-domain Visual Entity Recognition with Large-scale Knowledge Graphs via Contrastive Learning},
  author={Zhou, Hongkuan and Halilaj, Lavdim and Monka, Sebastian and Schmid, Stefan and Zhu, Yuqicheng and Wu, Jingcheng and Nazer, Nadeem and Staab, Steffen},
  journal={arXiv preprint arXiv:2510.13675},
  year={2025}
}
}


\appendix

\section{Overview of Appendixes}
In this supplementary material, we present implementation details and more experiments. 
First, we provide additional analysis in Section~\ref{sec:more_ablation}. The details of the training objective are provided in Section~\ref{sec:more_obj}. Then, we provide the retrieval results in Section~\ref{sec:ret_results} and the clarification on the reason for choosing the primary image in Section~\ref{sec:most_rep_image}. Finally, we provide a visualization in Section~\ref{sec:more_visualization}.



\subsection{More Ablation Study}
\label{sec:more_ablation}
\paragraph{The Impact of Training Iteration.}
As shown in Figure~\ref{fig:scale} and aligned with findings from OVEN, WikiCLIP achieves strong open-domain recognition efficiency: its unseen entity accuracy peaks at 4K iterations (with only 0.4M training samples) before slightly declining as seen entity accuracy continues to improve with extended training. This demonstrates an effective balance – while prolonged training introduces mild unseen entity performance degradation ($<$3\%), the framework maintains competitive results through early stopping, validating its data efficiency and practical viability for visual-entity retrieval tasks.

\begin{figure*}[ht]
    \centering
     \includegraphics[width=0.8\linewidth]{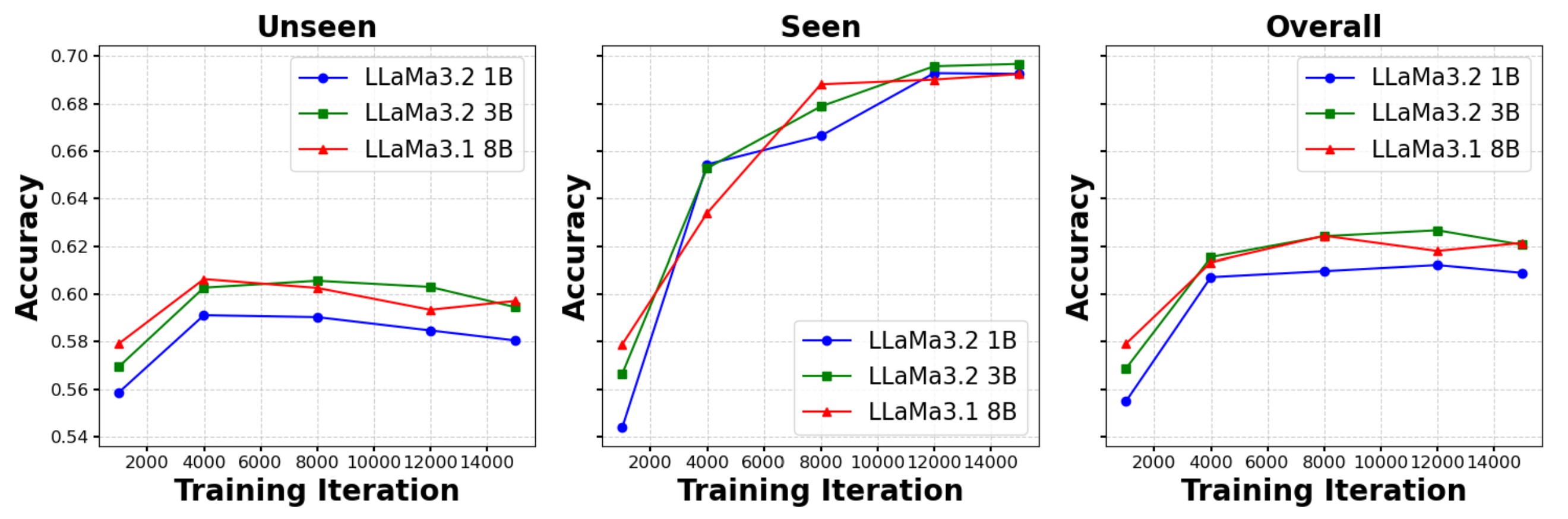}   
    \caption{\textbf{Performance with Varying Training Iterations and LLM Choices.} We report the accuracy of the INFOSEEK validation set of WikiCLIP using three different scales of LLMs, along with varying training iterations.
    }
    \label{fig:scale}
\end{figure*}



\paragraph{Text Encoder Ablation on OVEN and EVQA}
We provide the experiments about the choice of text encoder on OVEN and EVQA datasets in Tab~\ref{tab:r3_minor}, which requires the model to match the query from a 2M entity set instead of 100k for InfoSeek. We observe that switching to LLaMA3 leads to significant improvements, which indicates that providing rich semantic information helps entity discrimination.

\paragraph{LLM Capacity \& Text Length Joint Analysis}
To further investigate the interplay between LLM scale and input text length, we jointly vary these two factors in Table~\ref{tab:r2_major2}. The results are consistent with the findings in Sec~\ref{sec4.6} of the main paper: simply scaling up the LLM or extending the raw Wikipedia text does not lead to substantial gains, suggesting that future efforts should focus on extracting more informative content rather than increasing model capacity or input length alone.

\paragraph{Impact of Visual Encoder Scaling}
To further investigate the role of visual encoder scale, we reproduce the strongest contrastive baseline, CLIP2CLIP, using EVA-CLIP-8B. As shown in Table~\ref{tab:baseline_re}, scaling up the visual encoder alone does not allow the contrastive baseline to match the performance of WikiCLIP, suggesting that the knowledge-aware alignment mechanism provides benefits beyond encoder scaling.

\begin{table}[t]
\centering

\scriptsize
\begin{tabular}{c|ccc|ccc}
\hline
& \multicolumn{3}{c|}{EVQA} & \multicolumn{3}{c}{OVEN} \\
Model & Unseen & Seen & Overall & Unseen & Seen & HM \\
\hline
EVA-CLIP 8B & 12.9 & 12.7 & 14.1 & 17.3 & 15.6 & 15.6 \\
LLaMa3.2 1B & 27.7 & 39.9 & 30.7 & 27.0 & 36.8 & 31.1 \\
\hline
\end{tabular}
\caption{Comparison of text encoders on EVQA and OVEN.}
\label{tab:r3_minor}
\end{table}

\begin{table}[]
    \centering
    \scriptsize
        \begin{tabular}{c|ccc}
        \hline
        \textbf{Seen/Unseen/All} & \textbf{128}             & \textbf{256}            & \textbf{512}            \\ \hline
        LLAMA3.2-1B     & 66.9/57.4/59.8  & 69.3/58.5/61.2 & 67.4/57.8/60.2 \\
        LLAMA3.2-3B     & 66.5 /59.8/61.5 & \textbf{69.6}/60.3/\textbf{62.7} & 66.6/\textbf{58.9}/60.8 \\
        LLAMA3.2-8B     & \textbf{69.0}/\textbf{60.2}/\textbf{62.4}  & 68.7/\textbf{60.4}/62.4 & \textbf{68.8}/58.8/\textbf{61.3} \\ \hline
        \end{tabular}
    \caption{Joint analysis of text length and LLM scale on the INFOSEEK.}
    \label{tab:r2_major2}
\end{table}

\begin{table}[]
    \centering
    \small
        \begin{tabular}{c|ccc}
        \hline
         & \textbf{Unseen}  & \textbf{Seen} & \textbf{HM}  \\ \hline
        EVA-CLIP-8B I2I        & 11.8 & 8.8   & 10.1 \\
        CLIP2CLIP with EVA-CLIP-8B         & 14.6 & 11.1   & 12.6 \\
        WikiCLIP-S        & \textbf{27.0}  & \textbf{36.8}    & \textbf{31.1} \\ \hline
        \end{tabular}
    \caption{Comparison of contrastive baselines with scaled visual encoder on OVEN.}
    \label{tab:baseline_re}
\end{table}

\paragraph{VGKA Architecture Ablation}
We further ablate the VGKA architecture in terms of the number of layers and attention heads, as shown in Table~\ref{tab:r2_major3}. Performance remains stable across all variants, suggesting that VGKA is robust as long as it is sufficiently expressive to support visual-guided knowledge alignment.

\begin{table}[]
    \centering
    \small
        \begin{tabular}{c|ccc}
        \hline
        \textbf{Layers / Heads} & \textbf{Seen} & \textbf{Unseen} & \textbf{All}  \\ \hline
        2 / 16         & \textbf{69.3} &  \textbf{58.5}  & \textbf{61.2} \\
        8 / 16         & 68.2 & 58.0   & 60.6 \\
        16 / 32        & 66.4 & 56.5   & 59.1 \\ \hline
        \end{tabular}
    \caption{VGKA architecture ablation on the INFOSEEK.}
    \label{tab:r2_major3}
\end{table}

    
    

\subsection{Contrastive Training Objective}
\label{sec:more_obj}
We provide the details of the training objective in this section.
Given a mini-batch 
$\mathcal{B} = \big[(\mathbf{h}_i, \mathbf{v}_i)\big]_{i=1}^N$,
the positive pair for the $i$-th query is $(\mathbf{h}_i, \mathbf{v}_i)$, and the in-batch negatives are denoted as $\mathcal{B}_i^- = \{ \mathbf{v}_j \mid j \neq i \}$. After applying our hard negative synthesis strategy, the negative set is replaced with a harder set $\widetilde{\mathcal{B}}_i^{-}$, where each negative sample is substituted by a synthetic hard negative $\mathbf{\tilde{v}}_j$ whenever it exhibits higher similarity to the query:
\[
\widetilde{\mathcal{B}}_i^{-} =
\left\{
\begin{array}{ll}
\mathbf{\tilde{v}}_j, & \text{if } \mathrm{Sim}(\mathbf{h}_i, \mathbf{\tilde{v}}_j) > \mathrm{Sim}(\mathbf{h}_i, \mathbf{v}_j), \\[4pt]
\mathbf{v}_j, & \text{otherwise},
\end{array}
\right.
\quad \forall j \neq i.
\]
We optimize an InfoNCE loss with cosine similarity and temperature $\tau$, defined for each query $\mathbf{h}_i$ as:
\begin{equation}
\mathcal{L}_{\text{InfoNCE}}^{(i)}
= -\log\frac{\exp(s_{i,i}/\tau)}
{\sum_{v\in\{\mathbf{v}_i\}\cup\widetilde{\mathcal{B}}_i^{-}}\exp(s_{i,v}/\tau)}.
\label{eq:infonce_compact}
\end{equation}

\noindent where
\[
s_{i,v} \triangleq \mathrm{Sim}(\mathbf{h}_i, \mathbf{v}).
\]
The final training loss is averaged across all samples in the mini-batch:
\[
\mathcal{L}_{\text{InfoNCE}} = \frac{1}{N} \sum_{i=1}^{N} \mathcal{L}_{\text{InfoNCE}}^{(i)}.
\]

\begin{table*}[]
\centering

\resizebox{0.8\linewidth}{!}{
\begin{tabular}{lcccccccccccc}
\toprule
\multirow{2}{*}{Methods} & \multicolumn{4}{c}{HM@K (OVEN)} & \multicolumn{4}{c}{Recall@K (EVQA)} & \multicolumn{4}{c}{Recall@K (InfoSeek)} \\
\cmidrule(lr){2-5} \cmidrule(lr){6-9} \cmidrule(lr){10-13}
& K=1 & K=5 & K=10 & K=20 & K=1 & K=5 & K=10 & K=20 & K=1 & K=5 & K=10 & K=20 \\
\midrule
CLIP I-I       & 10.1 & 27.1 & 36.0 & 44.1 & 13.3 & 31.3 & 41.0 & 48.8 & 45.6 & 67.1 & 73.0 & 77.9 \\
CLIP I-T       & -    & -    & -    & -    & 3.3  & 7.7  & 12.1 & 16.5 & 32.0 & 54.0 & 61.6 & 68.2 \\
Google Lens    & -    & -    & -    & -    & 47.4 & 62.5 & 64.7 & 65.2 & -    & -    & -    & -    \\
Echosight      & -    & -    & -    & -    & 36.5 & 47.9 & 48.8 & 48.8 & 53.2 & 74.0 & 77.4 & 77.9 \\
WikiCLIP-S     & 31.1 & 53.7 & 61.3 & 67.8 & 30.7 & 53.6 & 62.5 & 69.1 & 61.2 & 76.8 & 81.8 & 86.4 \\
WikiCLIP-L     & 31.6 & 53.3 & 60.5 & 67.4 & 31.9 & 53.3 & 61.2 & 69.0 & 62.7 & 77.7 & 82.2 & 86.6 \\
\bottomrule
\end{tabular}}
\caption{\textbf{Performance comparison on open-domain VER benchmarks. HM@K for OVEN, Recall@K for EVQA and InfoSeek.}}
\label{tab:ver_results}
\end{table*}

\subsection{Topk Retrieval Results}
\label{sec:ret_results}
Table~\ref{tab:ver_results} presents the performance comparison of WikiCLIP and baseline methods on open-domain visual entity recognition (VER) tasks. WikiCLIP formulates VER as a retrieval task, providing top-$k$ retrieval results to evaluate its effectiveness. As shown in the table, WikiCLIP achieves the highest HM@20 score of 67.8 on the OVEN benchmark, significantly outperforming CLIP I-I. On EVQA, our method surpasses Google Lens at $K=20$, demonstrating that WikiCLIP can achieve surprisingly strong performance even with limited training data. Finally, on the InfoSeek benchmark, WikiCLIP achieves a recall@20 of 86.6, highlighting its effectiveness in real-world applications. These results validate the robustness and practicality of our approach for large-scale VER tasks.

\subsection{How to choose the most representative entity image?}
\label{sec:most_rep_image}
In the Vision-guided Knowledge Adaptor module, we use the most representative entity image to provide guidance to obtain the knowledge-aware entity representations. In practice, we use the lead image on the Wikipedia page as the most representative image. This selection is based on Wikipedia's content policies: lead images should be natural and appropriate representations of the entity (in accordance with \textit{Wikipedia:Manual of Style/Images}).

\subsection{Visualization}
\label{sec:more_visualization}
\begin{figure*}[]
    \centering
     \includegraphics[width=1.0\linewidth]{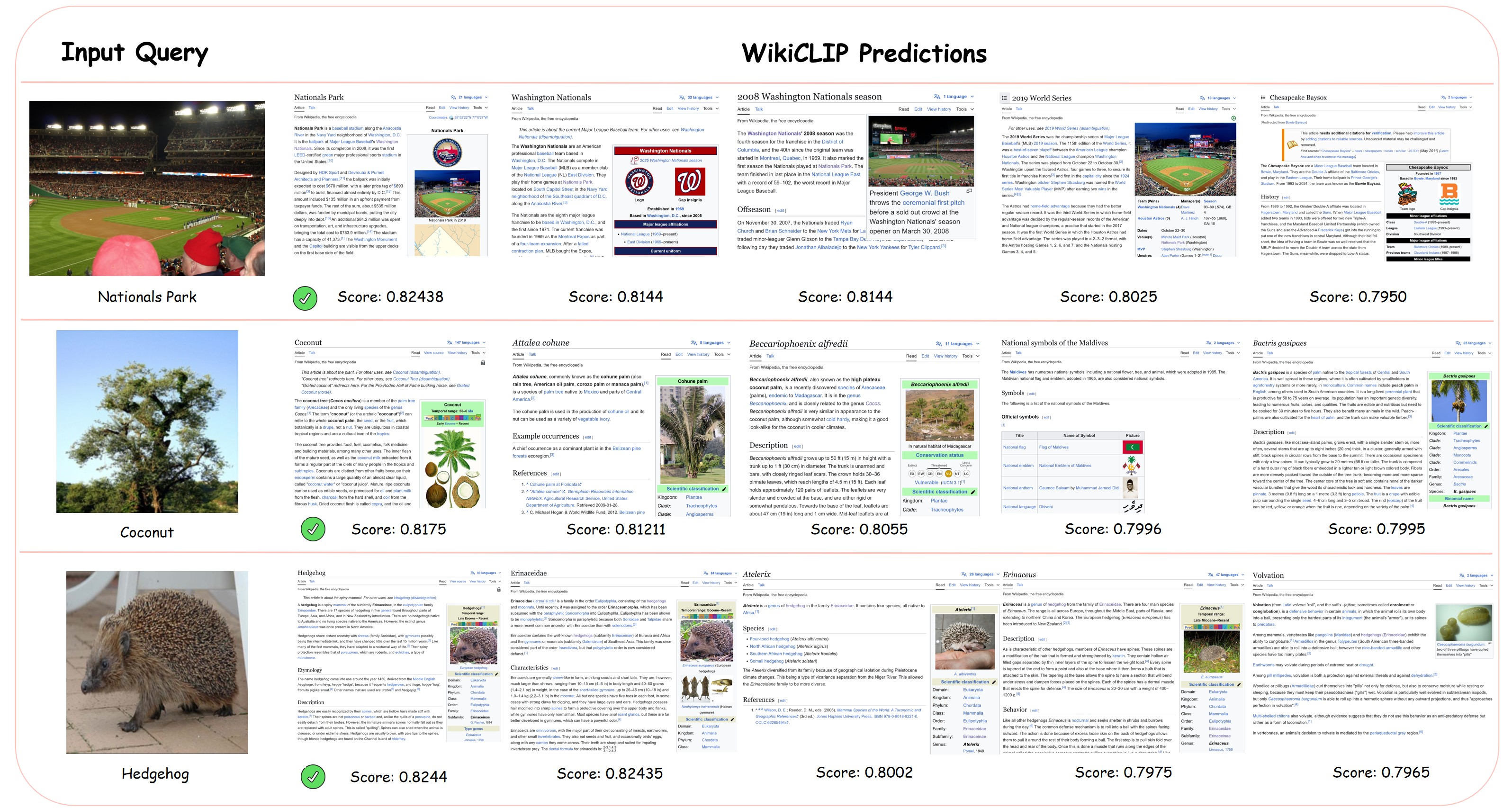}   
    \caption{\textbf{The Visualization of Topk prediction of WikiCLIP.}}
    \label{fig:vis}
\end{figure*}

\paragraph{Retrieval Results Visualization}
To provide an intuitive understanding of WikiCLIP’s effectiveness, we present qualitative visualization samples in Figure~\ref{fig:vis}. The figure showcases the top-5 retrieval results for various queries, illustrating how WikiCLIP performs in visually ambiguous cases. As observed, the most challenging samples often involve entities with highly similar visual features, making fine-grained discrimination difficult. However, WikiCLIP successfully retrieves the correct ground-truth entity by leveraging textual descriptions, demonstrating its ability to incorporate semantic knowledge for precise entity recognition. These visualizations highlight the strength of our method in resolving challenging cases where purely visual matching would struggle.

\begin{figure*}[]
    \centering
     \includegraphics[width=1.0\linewidth]{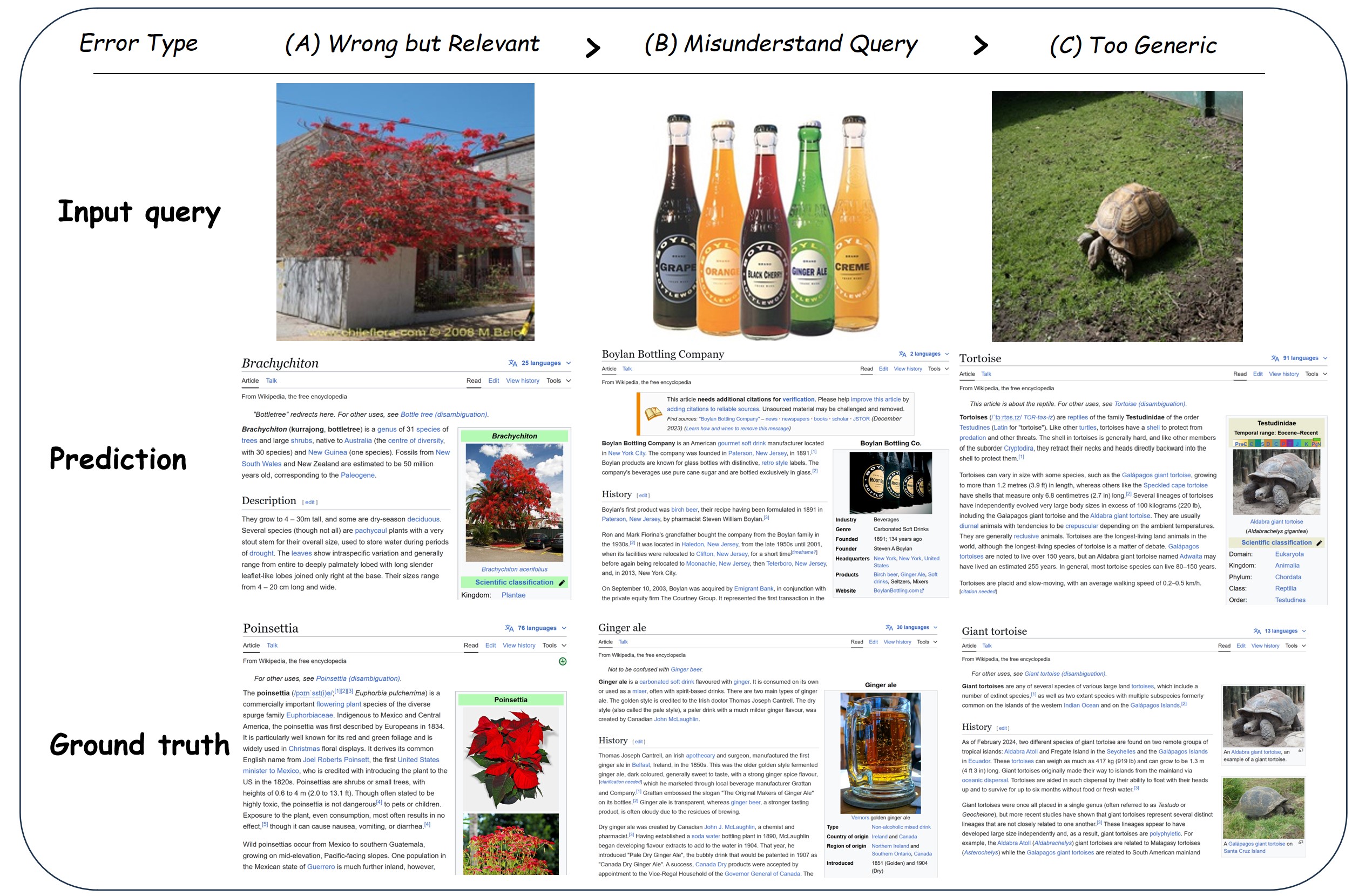}   
    \caption{\textbf{The Visualization of Error Case of WikiCLIP.}}
    \label{fig:error}
\end{figure*}

\begin{figure*}[t]
    \centering
    \begin{subfigure}[b]{0.9\linewidth}
        \includegraphics[width=\linewidth]{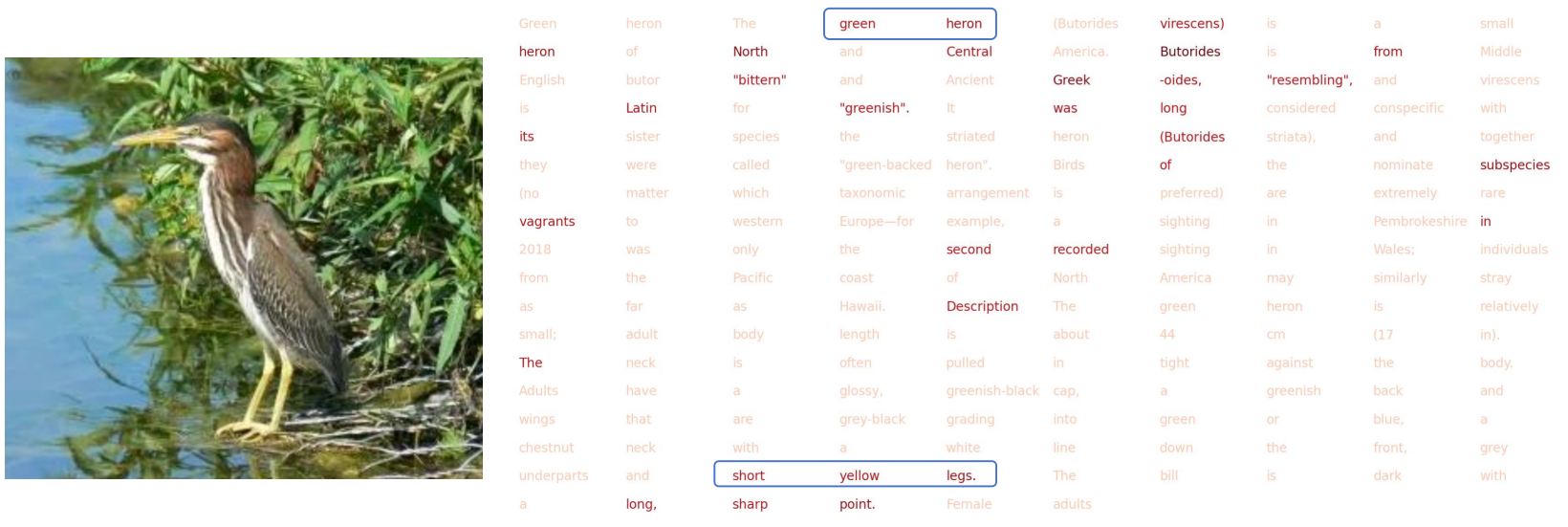}
        \caption{}
        \label{subfig:attn_vis1}
    \end{subfigure}
    
    \begin{subfigure}[b]{0.9\linewidth}
        \includegraphics[width=\linewidth]{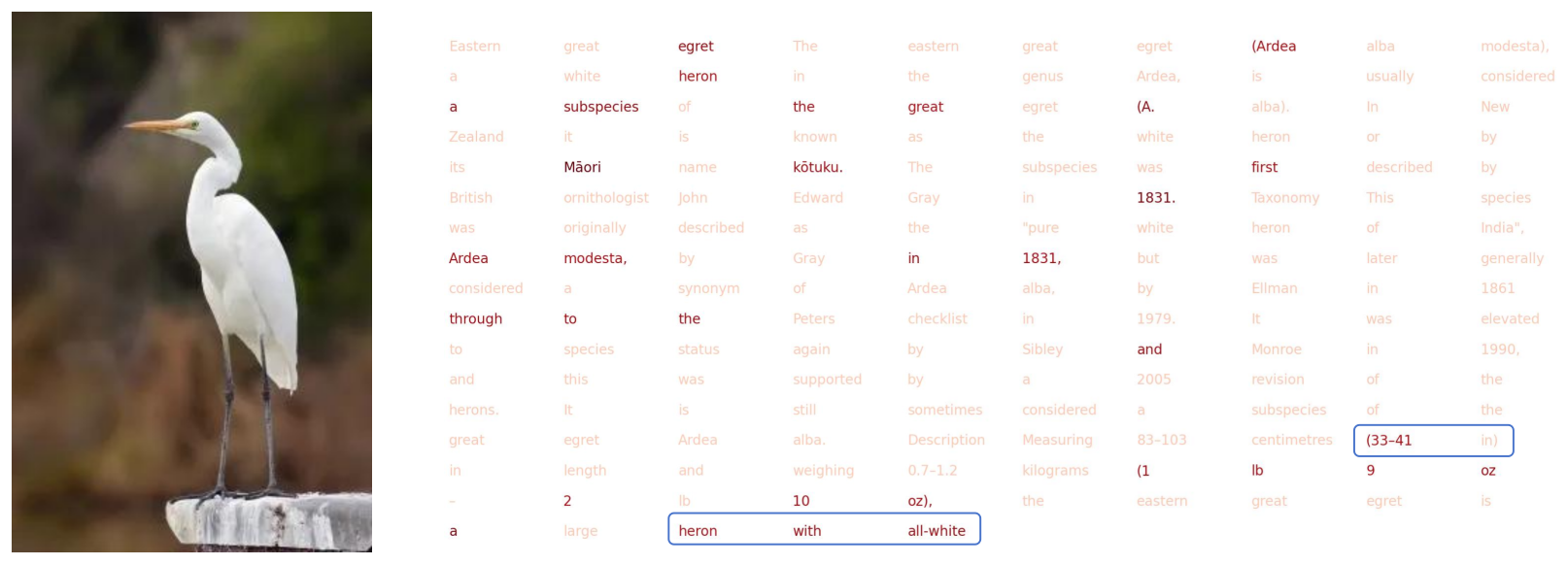}
        \caption{}
        \label{subfig:attn_vis2}
    \end{subfigure}
    
    \begin{subfigure}[b]{0.9\linewidth}
        \includegraphics[width=\linewidth]{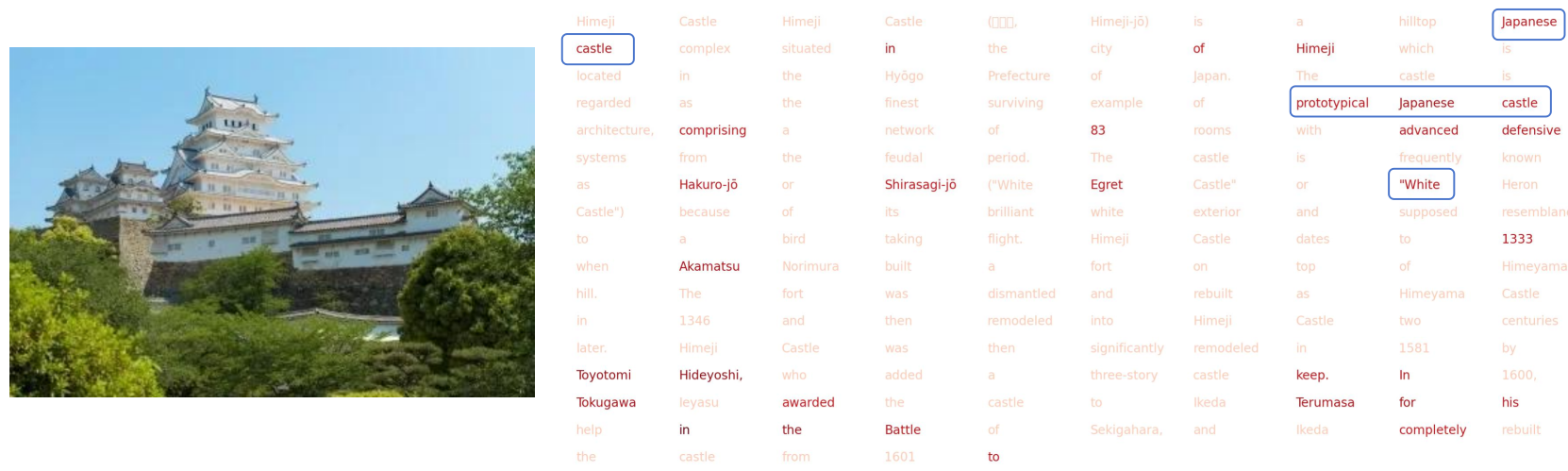}
        \caption{}
        \label{subfig:attn_vis3}
    \end{subfigure}
    
    \caption{\textbf{Vision-guided Knowledge Selection Visualization.}}
    \label{fig:attn_vis}
\end{figure*}

\paragraph{Vision-guided Knowledge Selection Visualization}
To provide an intuitive understanding of the Vision-guided Knowledge Selection module, we present qualitative visualization samples in Figure~\ref{fig:attn_vis}. Specifically, we show the attention map of each text token when guided by patch-level vision signals. For the sake of clarity, we sum the attention from each image patch and highlight the top 32 text segments with higher attention. We circle the text segments that we believe are helpful for entity discrimination.
As observed, the Vision-guided Knowledge Selection is able to detect the detailed discriminative features of entities.

\paragraph{Error Case Analysis}
To better understand the limitations of WikiCLIP, we analyzed its prediction errors on OVEN and identified three main types of failure cases, as illustrated in Figure~\ref{fig:error}. The most prevalent error type is Wrong but Relevant, where the predicted entity is semantically related to the ground truth but still incorrect. While our method leverages textual information to mitigate such errors, open-domain visual entity recognition remains a highly challenging task.
The second type of error arises when the ground truth entity is not directly related to any entity present in the image. We attribute this to annotation noise in OVEN’s entity split and query split, which affects model performance.
The final category of errors pertains to prediction granularity, where the predicted entity is at an incorrect level of specificity. We believe this issue is also a result of annotation inconsistencies in OVEN. Due to the inherent noise in dataset annotations, there is no straightforward solution to fully address this problem.

\begin{figure*}[t]
    \centering
    
    \begin{subfigure}{0.44\textwidth}
        \includegraphics[width=\linewidth]{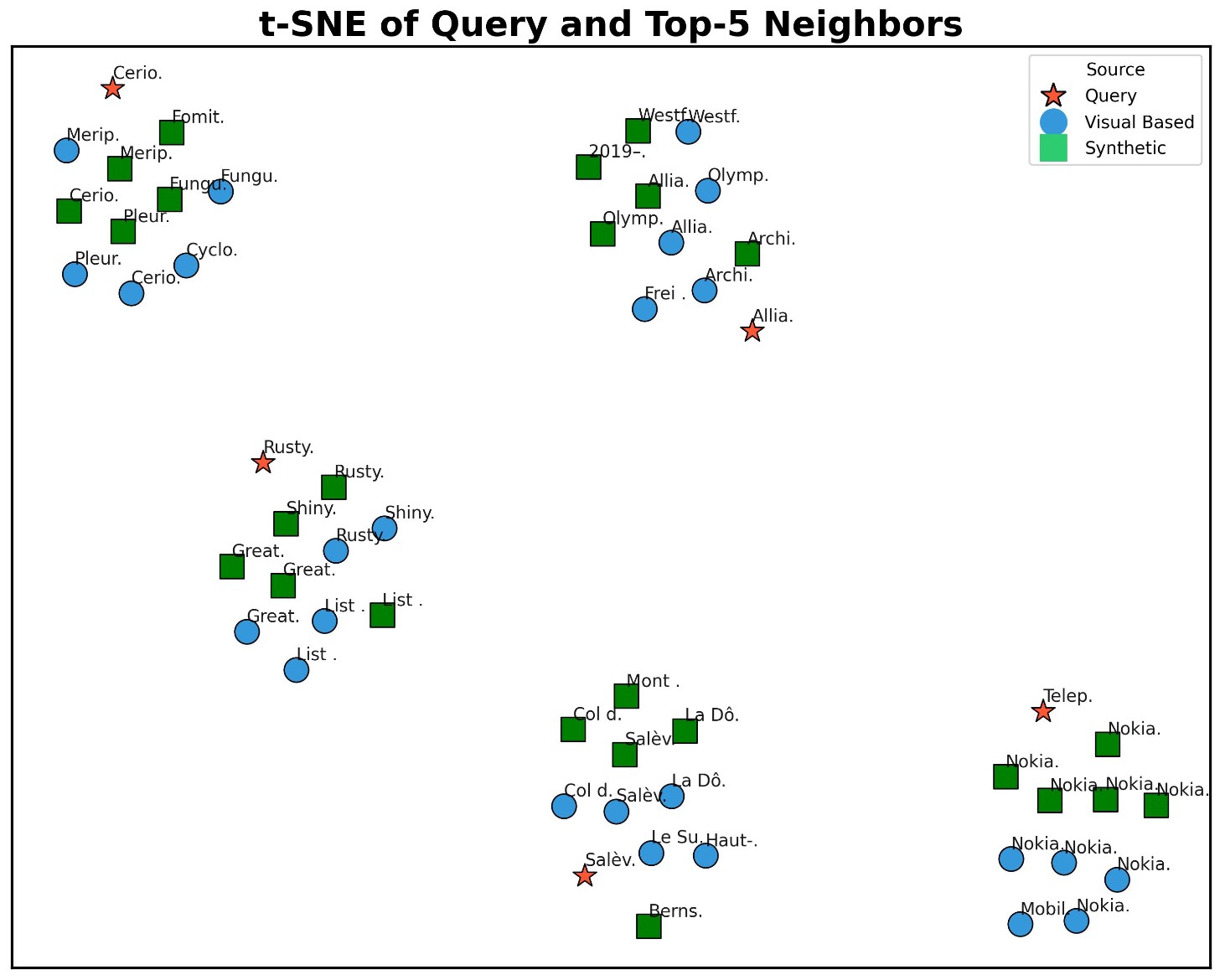}
        \caption{}
         \label{fig:avf_performance}
        \end{subfigure}
        \hfill
        \begin{subfigure}{0.54\textwidth}
        \includegraphics[width=\linewidth]{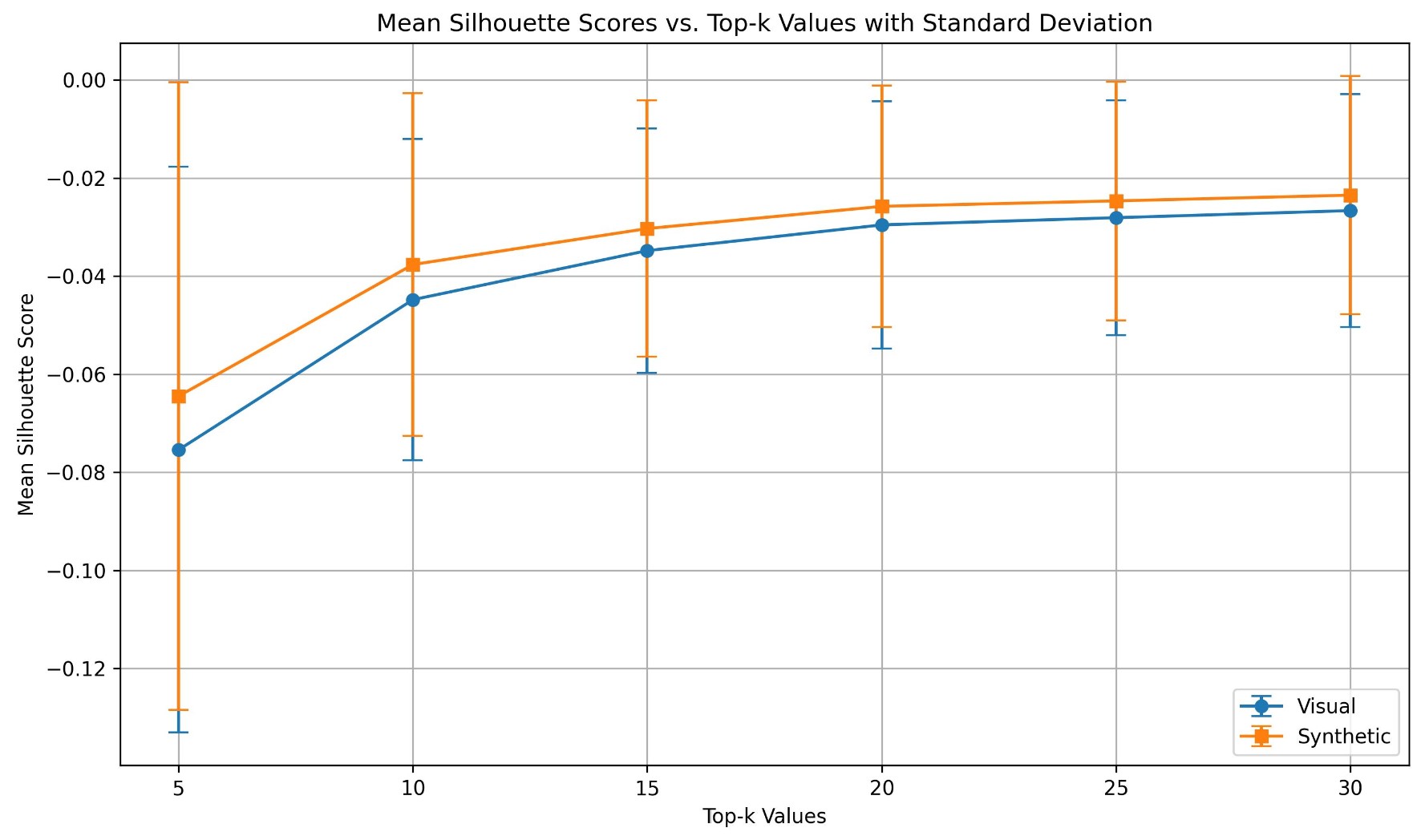}
            \caption{}
            \label{fig:easy_hard}
    \end{subfigure}
\caption{\textbf{Visualization of Hard Negatives.}}
\label{fig:vis_hard}
\end{figure*}

    
    
    

\paragraph{Visualization of Hard Negatives.}
To intuitively demonstrate the effectiveness of our hard negative synthesis strategy, we visualize the distribution of entity representations. Specifically, we randomly sample a query image and retrieve the top-5 closest entity representations with and without hard negative synthesis. As shown in Figure~\ref{fig:vis_hard}, we apply t-SNE to project these representations into a 2D space. It can be observed that the entity representations trained with hard negative synthesis exhibit a more sparse and discriminative structure across different classes.

To quantitatively validate this observation, we compute the Silhouette Score, a widely-used metric that evaluates how similar a sample is to its own cluster (cohesion) compared to other clusters (separation). A higher Silhouette Score indicates more compact intra-class representations and better-separated inter-class representations. As shown in Figure~\ref{fig:vis_hard}, hard negative synthesis consistently leads to higher Silhouette Scores, confirming that it encourages more structured and distinguishable entity representation spaces.

\end{document}